\documentclass[10pt,journal,cspaper,compsoc]{IEEEtran}

\usepackage{amsthm}
\usepackage{graphicx}
\usepackage{supertabular}
\usepackage{slashbox}
\usepackage{subfigure}
\usepackage{rotating}
\usepackage{amssymb}
\usepackage{bm}

\usepackage{stmaryrd}
\usepackage{amsfonts}
\usepackage{array}
\usepackage{amssymb}
\usepackage{algorithmic}
\usepackage{algorithm}
\usepackage{multirow}
\usepackage[cmex10]{amsmath}
\usepackage{bm}
\usepackage{cite}
\usepackage{color}
\usepackage{threeparttable}
\usepackage{enumerate}
\usepackage{mathrsfs}

\usepackage{tabularx}

\usepackage{setspace}

\usepackage{graphics}
\usepackage{graphicx}
\usepackage{epsfig}

\newtheorem{lemma}{Lemma}
\newtheorem{proposition}{Proposition}
\newtheorem{definition}{Definition}
\newtheorem{theorem}{Theorem}

\newtheorem{example}{Example}

\usepackage{url}

\begin{document}


%
\title{Unsupervised Ranking of Multi-Attribute Objects Based on Principal Curves}
\author{Chun-Guo~Li, Xing~Mei,
    and~Bao-Gang~Hu,~\IEEEmembership{Senior Member,~IEEE,}
\IEEEcompsocitemizethanks{\IEEEcompsocthanksitem C.-G. Li, X. Mei and B.-G. Hu
are with the National Laboratory of Pattern Recognition,
Institute of Automation Chinese Academy of Sciences, 95 ZhongGuanCun East Road, Beijing
100190, P.R. China. \protect\\
Email: cgli@nlpr.ia.ac.cn, hubg@nlpr.ia.ac.cn
\IEEEcompsocthanksitem C.G. Li is also with Faculty of Mathematics and Computer Science, Hebei
University, 180 Wusi East Road, Baoding, Hebei 071002, P.R. China.}
\thanks{}}

\IEEEcompsoctitleabstractindextext{%

\begin{abstract}
Unsupervised ranking faces one critical challenge in evaluation
applications, that is, no ground truth is available. When PageRank
and its variants show a good solution in related subjects, they are
applicable only for ranking from link-structure data. In this work,
we focus on unsupervised ranking from multi-attribute data which is
also common in evaluation tasks. To overcome the challenge, we
propose five essential meta-rules for the design and assessment of
unsupervised ranking approaches: \emph{scale and translation
invariance}, \emph{strict monotonicity}, \emph{linear/nonlinear
capacities}, \emph{smoothness}, and \emph{explicitness of parameter
size}. These meta-rules are regarded as high level knowledge for
unsupervised ranking tasks. Inspired by the works in
\cite{gorban2010b} and \cite{hubg1998}, we propose a ranking
principal curve (RPC) model, which learns a one-dimensional manifold
function to perform unsupervised ranking tasks on multi-attribute
observations. Furthermore, the RPC is modeled to be a cubic
B\'{e}zier curve with control points restricted in the interior of a
hypercube, thereby complying with all the five meta-rules to infer a
reasonable ranking list. With control points as the model
parameters, one is able to understand the learned manifold and to
interpret the ranking list semantically. Numerical experiments of
the presented RPC model are conducted on two open datasets of
different ranking applications. In comparison with the
state-of-the-art approaches, the new model is able to show more
reasonable ranking lists.
\end{abstract}

}

\maketitle

\IEEEdisplaynotcompsoctitleabstractindextext

\IEEEpeerreviewmaketitle

\section{Introduction}\label{sec:intro}

\begin{figure*}[!htbp]
\centering
\includegraphics[scale=0.5, bb = 0 0 768 268]{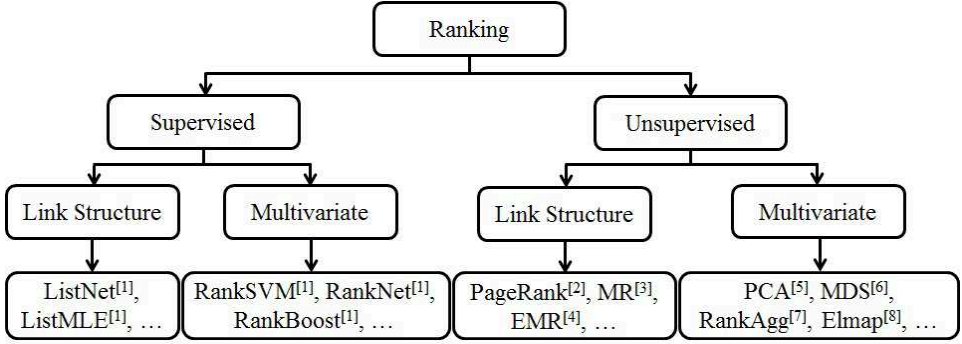}
\caption{Hierarchical diagram of ranking approaches. RPC is an
unsupervised ranking approach based on multi-attribute observations
for objects. } \label{fig:structure}
\end{figure*}

\IEEEPARstart{F}{rom}
the viewpoint of machine learning, ranking can be performed in an either supervised or unsupervised way
as shown in the hierarchical structure in Fig. \ref{fig:structure}.
When supervised ranking \cite{lihang2011} is able to evaluate the ranking performance from the given
ground truth, unsupervised ranking seems more challenging because no ground truth
label is available.
Modelers or users will encounter a more difficult issue below:

\hangafter=1 \setlength{\hangindent}{1.5em}
``\emph{How can we insure that the ranking list from the unsupervised
ranking is reasonable or proper?}''

\noindent From the viewpoint of given data types, ranking approaches
can be further divided into two categories: ranking based on link structure and
ranking based on multi-attribute data. PageRank \cite{page1998} is
one of the representative unsupervised approaches to rank items
which have a linking network (e.g. websites). But PageRank and its
variants do not work for ranking candidates which have no links. In
this paper, we focus on unsupervised ranking approaches on a set of
objects with multi-attribute numerical observations.

To rank from multi-attribute objects, weighted summation of
attributes is widely used to provide a scalar score for each object.
But different weight assignments give different ranking lists such
that ranking results are not convincing enough. The first principal
component analysis (PCA) provides a weight learning approach
\cite{bishop2006}, by which the score for each object is determined
by its principal component on the skeleton of the data distribution.
However, it encounters problems when the data distribution is
nonlinearly shaped. Although kernel PCA \cite{bishop2006} is
proposed to attack this problem, the mapping to the kernel space is
not order-preserving, which is the basic requirement for a ranking
function. Neither dimension reduction methods \cite{guyon2003}
nor vector quantization \cite{vasuki2006} can assign scores
for multi-attribute observations.

\begin{figure}[t]
\centering \mbox{ \subfigure[Polyline Approximation (non-strict
monotonicity)]{\includegraphics[scale = 0.3, bb = 0 0 341
267]{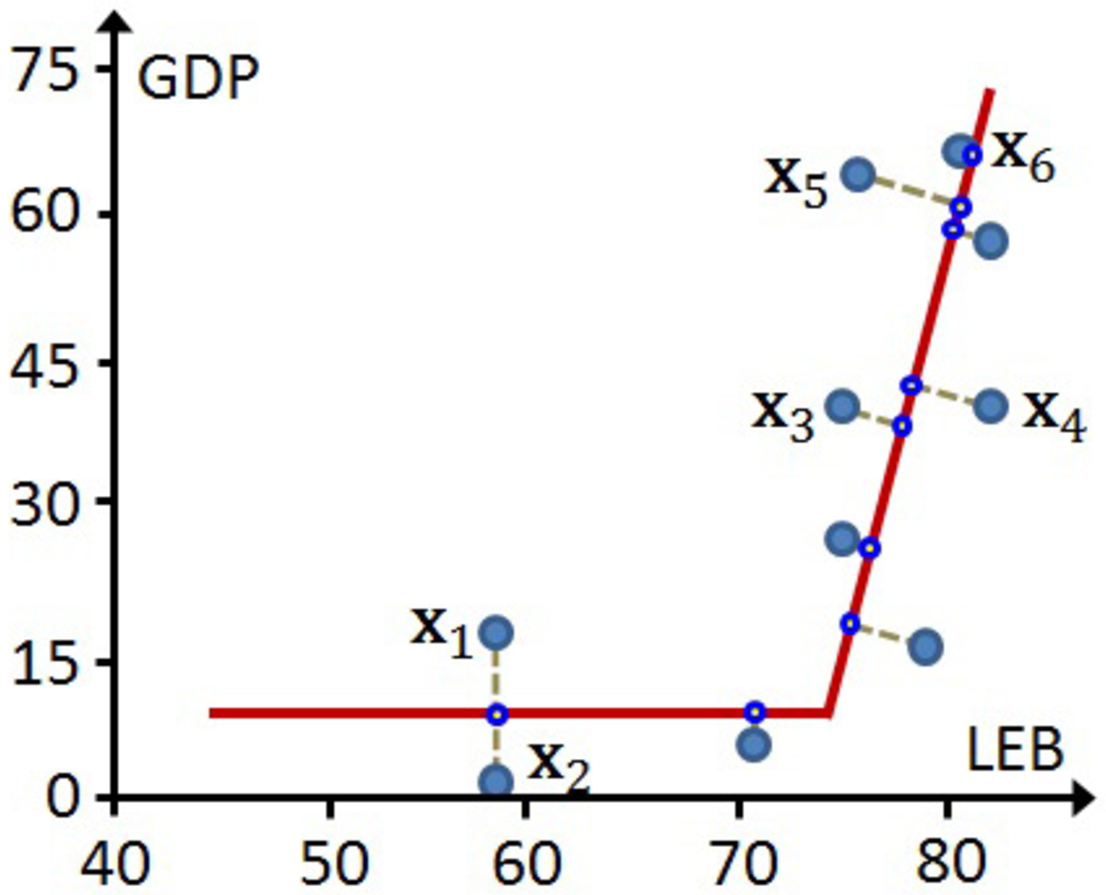}} \hspace{2em} \subfigure[A General Principal
Curve (non-monotonicity)]{\includegraphics[scale = 0.3, bb = 0 0 341
267]{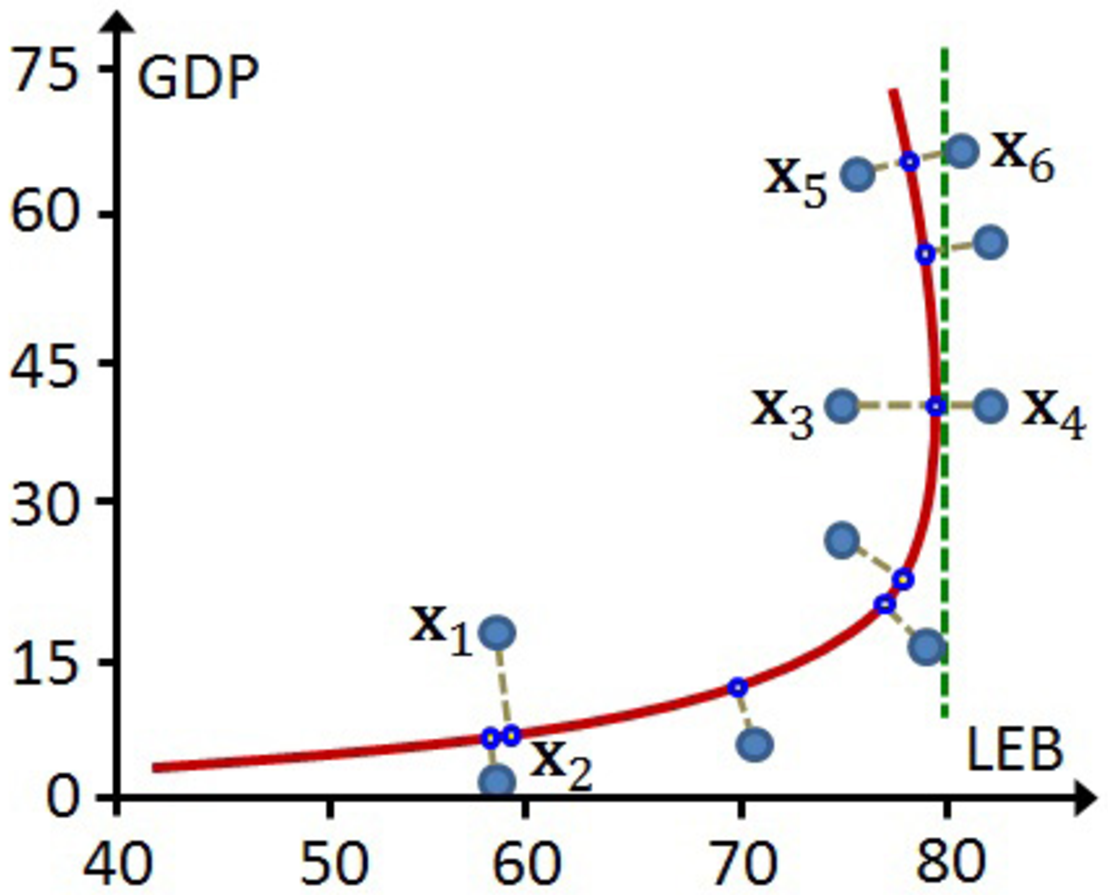}} } \caption{Examples on a monotonicity property
for ranking with principal curves. } \label{fig:drawback}
\end{figure}

As the nonlinear extension of the first PCA, principal curves can be
used to perform a ranking task \cite{hastie1989,gorban2010b}. A
principal curve provides an ordering of data points by the ordering
of threading through their projected points on the curve
(illustrated by Fig. \ref{fig:drawback}) which can be regarded as
the ``ranking skeleton''. However, not all of principal curve models
are capable of performing a ranking task. Polyline approximation of a
principal curve \cite{kegl2000} fails to provide a consistent
ranking rule due to non-smoothness at connecting points. Besides, it
fails to guarantee order-preserving. Order-preserving can not be
guaranteed either by a general principal curve model (e.g.
\cite{gorban2010}) which is not modeled specially for ranking tasks.
The problem can be tackled by the constraint of strict monotonicity
which is one of the constraints we present for ranking functions in
this paper. Example \ref{emp:gdp2leb} shows that strict monotonicity is a
necessary condition for a ranking function but was neglected by all
other investigations.

\begin{example}\label{emp:gdp2leb}
 \emph{Suppose we want to evaluate life qualities of countries
 with a principal curve based on two attributes:
 LEB\footnote{Life Expectancy at Birth, years}
 and GDP\footnote{Gross Domestic Product per capita by Purchasing Power Parities, K\textdollar/person}.
 Each country is a data point in the two-dimensional plane of LEB and GDP.
 If the principal curve is approximated by a polyline as in Fig. \ref{fig:drawback}(a),
 the piece of the horizontal line is not strictly monotone.
 It makes the same ranking
 solution for $\mathbf{x}_1=(58, 1.4)$ and $\mathbf{x}_2=(58, 16.2)$
 but $\mathbf{x}_2$ should be ranked higher than $\mathbf{x}_1$.
 For a general principal curve like the curve in Fig. \ref{fig:drawback}(b)
 which is not monotone, two pairs of points are ordered unreasonably.
 The pair, $\mathbf{x}_3=(74, 40.2)$ and $\mathbf{x}_4=(82,
 40.2)$, are put in the same place of the ranking list
 since they are projected to the same point which has the vertical
 tangent line to the curve.
 But $\mathbf{x}_4$ should be ranked higher for its higher LEB than $\mathbf{x}_3$.
 Another pair, $\mathbf{x}_5=(75, 62.5)$ and $\mathbf{x}_6=(81,
 64.8)$, are also put in the same place but apparently $\mathbf{x}_6$
 should be ranked higher than $\mathbf{x}_5$.
 With strict monotonicity, these points would be in the order that
 they are.
 }
\end{example}

\begin{figure}[t]
\centering
\includegraphics[scale=0.4, bb = 0 0 410 214]{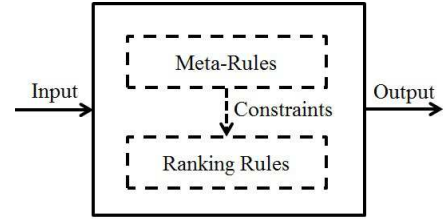}
\caption{Motivation of RPC model for unsupervised ranking.}
\label{fig:rpcmodel}
\end{figure}

Following the principle of ``\textit{let the data speak for themselves}'' \cite{gould1981},
this work tries to attack problems for unsupervised ranking of multi-attribute objects
with principal curves .
First, ranking performance is taken into account for the design of ranking functions.
It is known that knowledge of a given task can always improve learning performance \cite{hubg2009}.
The reason why PageRank produces a commonly acceptable search result for a query, lies on
that PageRank algorithm is designed by integrating the knowledge about backlinks \cite{page1998}.
For multi-attribute objects with no linking networks, knowledge about ranking functions
can be taken into account to make ranking functions produce reasonable ranking lists.
In this work, we present five essential meta-rules for ranking rules (Fig. \ref{fig:rpcmodel}).
These meta-rules can be capable of assessing the reasonability of ranking lists
for unsupervised ranking.

Second, principal curves should be modeled to be able to serve as ranking functions.
As referred in \cite{gorban2010b}, ranking with a principal curve is performed on the
learned skeleton of data distribution.
But not all principal curve models are capable of producing reasonable ranked lists
when no ranking knowledge is embedded into principal curve models.
Motivated by \cite{hubg1998}, the principal curve can be parametrically designed with
a cubic B\'{e}zier curve.
We will show in Section \ref{sec:rpc} that the parameterized principal curve has all the five
meta-rules with constraints on control points and that its existence and convergency
of learning algorithm are proved theoretically.
Therefore, the parameterized principal curve is capable of making a reasonable ranking list.

The following points highlight the main contributions of this paper:
\begin{itemize}
 \item We propose five meta-rules for unsupervised ranking, which serve as
  high-level guidance in the design and assessment of unsupervised ranking approaches
  for multi-attribute objects. We justify that the five meta-rules are essential in
  applications, but unfortunately some or all of them were overlooked by most of ranking approaches.
 \item A \textit{ranking principal curve} (RPC) model is presented for unsupervised ranking from
  multi-attribute numerical observations of objects, different from PageRank
  which ranks from link structure \cite{page1998}.
  The presented model can satisfy all of five meta-rules for ranking tasks,
  while other existing approaches \cite{gorban2010b} overlooked them.
 \item We develop the RPC learning algorithm, and theoretically prove the existence of a RPC
  and convergency of learning algorithm for given multi-attribute objects for ranking.
  With RPC learning algorithm, reasonable ranking lists for openly accessible data illustrate
  the good performance of the proposed unsupervised ranking approaches.
\end{itemize}

\subsection{Related Works}

Domain knowledge can be integrated into leaning models to improve learning performance.
By coupling domain knowledge as prior information with network constructions,
Hu et al. \cite{hubg2009} and Daniels et al. \cite{daniels2010tnn} improve the prediction accuracy
of neural networks.
Recently, monotonicity is taken into consideration as constraints by Kot\l{}owski
et al. \cite{kotlow2013} to improved the ordinal classification performance.
For unsupervised ranking, the domain knowledge of monotonicity can also be taken into account
and is capable of assessing the ranking performance,
other than evaluation of side-effects \cite{pei2014tkde}.

Ranking on manifolds has provided a new ranking framework \cite{zhou2004,gorban2010b,cai2011,cheng2013},
which is different from general ranking functions such as ranking aggregation \cite{small2009}.
As one-dimensional manifolds, principal curves are able to perform unsupervised ranking tasks
from multi-attribute numerical observations of objects \cite{gorban2010b}.
But not all principal curve models can serve as ranking functions.
For example, Elmap can well portray the contour of a molecular surface \cite{gorban2010}
but would bring about a biased ranking list due to no guarantee of order-preserving \cite{gorban2010b}.
What's more, Elmap is hardly interpretable since the parameter size of principal curves
is unknown explicitly.

\begin{figure}[!t]
\centering
\includegraphics[width=3in, height=3in, bb = 0 0 590 590]{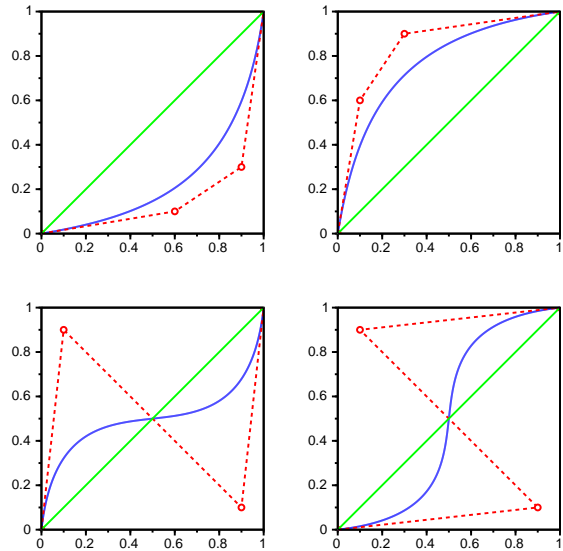}
\caption{For an increasing monotone function,
there are four basic nonlinear shapes \cite{hubg1998} of cubic
B\'{e}zier curves (in blue) which mimic shapes of the control polylines (in red).
Curve shapes are determined by the locations of control points. }
\label{fig:bcshape}
\end{figure}

A B\'{e}zier curve is a parametrical one-dimensional curve which is
widely used in fitting \cite{pastva1998}. Hu et al. \cite{hubg1998}
proved that in two-dimensional space a cubic B\'{e}zier curve is
strictly monotone with \textit{end points} in the opposite corner
and \textit{control points} in the interior of the square box as
shown in Fig. \ref{fig:bcshape}. To avoid confusion, end points
refer to the points on both ends of the control polyline (also the
end points of the curve) and control points refer to the other
vertices of the control polyline in this paper.

\subsection{Paper Organization}

The rest of this paper is organized as follows.
Backgrounds of this paper are formalized in the next section.
In Section \ref{sec:metarule}, five meta-rules are elaborated
for ranking functions.
In Section \ref{sec:rpc}, a ranking model, namely ranking principal curve (RPC) model,
is defined and formulated with a cubic B\`{e}zier curve which is proved to follow all the five
meta-rules for ranking functions.
RPC learning algorithm is designed to learn the control points of the cubic B\`{e}zier curve
in Section \ref{sec:rpclearning}.
To illustrate the effective performance of the proposed RPC model,
applications on real world datasets are carried out in Section \ref{sec:applications},
prior to summary of this paper in Section \ref{sec:conclu}.

\section{Backgrounds} \label{sec:related}

Consider ranking a set of $n$ objects
$\mathscr{A}=\{a_1,a_2,\cdots,a_n\}$ according to $d$ real-valued
attributes (or indicators, features)
$\mathbf{V}=\{v_1,v_2,\cdots,v_d\}$. Numerical observations of one
object $a\in\mathscr{A}$ on all the attributes comprise an item
which is denoted as a vector $\mathbf{x}$ in $d$-dimensional space
$\mathbf{R}^d$. Ranking objects in $\mathscr{A}$ is equivalent to
ranking data points
$\mathbf{X}=\{\mathbf{x}_1,\mathbf{x}_2,\cdots,\mathbf{x}_n\}$. That
is, to give the ordering of
$a_{i_1}\preceq{a_{i_2}}\preceq\cdots\preceq{a_{i_n}}$ can be
achieved by discovering the ordering of
$\mathbf{x}_{i_1}\preceq{\mathbf{x}_{i_2}}\preceq\cdots\preceq{\mathbf{x}_{i_n}}$
where $\{i_1,i_2,\cdots,i_n\}$ is a permutation of
$\{1,2,\cdots,n\}$ and $\mathbf{x}_i\preceq{\mathbf{x}_j}$ means
that $\mathbf{x}_i$ precedes $\mathbf{x}_j$. As there is no label to
help with ranking, it is an unsupervised ranking problem from
multi-attribute data.

Mathematically, ranking task is to provide a list of totally ordered points.
A total order is a special partial order which requires
comparability in addition to the requirements of reflexivity, antisymmetry and
transitivity for the partial order \cite{boyd2004}.
Let $\mathbf{x}$ and $\mathbf{y}$ are one pair of points in $\mathbf{X}$.
For ranking, if $\mathbf{x}$ and $\mathbf{y}$ are different, they have the ordinal
relation of either $\mathbf{x}\preceq\mathbf{y}$ or $\mathbf{y}\preceq\mathbf{x}$.
If $\mathbf{x}\preceq\mathbf{y}$ and $\mathbf{y}\preceq\mathbf{x}$, then $\mathbf{y}=\mathbf{x}$
which infers that $\mathbf{x}$ and $\mathbf{y}$ are the same thing.

Remembering that a partial order is associated with a proper cone and that
$\mathbf{R}^d_+$ is a \textit{self-dual} proper cone \cite{boyd2004}
$\mathbf{R}^d_+=\{{\bm\rho}:{\bm\rho}^T\mathbf{x}\geq0,\forall\mathbf{x}\in\mathbf{R}^d_+\}$,
the \textit{order} for ranking tasks on $\mathbf{R}^d$ is defined in this paper to be
\begin{equation}\label{eq:orderDef}
 \mathbf{x}\preceq\mathbf{y} \Longleftrightarrow
\left(\begin{array}{c}
    \delta_1(y_1-x_1) \\
    \delta_2(y_2-x_2) \\
        \vdots \\
    \delta_d(y_d-x_d)
       \end{array} \right)\in\mathbf{R}^d_+
\end{equation}
where $\mathbf{x}=(x_1,x_2,\cdots,x_d)^T$, $\mathbf{y}=(y_1,y_2,\cdots,y_d)^T$, and
\begin{equation} \label{eq:monoflag}
 \delta_j=\left\{\begin{array}{rl}
                  1, & j\in\mathbf{E}\\
         -1, & j\in\mathbf{F}
                 \end{array}\right. .
\end{equation}
It is easy to verify that the order defined by Eq.(\ref{eq:orderDef}) is a total
order with properties of comparability, reflexivity, antisymmetry and transitivity.
In Eq.(\ref{eq:monoflag}),
$\mathbf{E}$ and $\mathbf{F}$ are two subsets of $\{1,2,\cdots,d\}$ such that
$\mathbf{E}\bigcup\mathbf{F}=\{1,2,\cdots,d\}$ and
$\mathbf{E}\bigcap\mathbf{F}=\varnothing$.
If let
\begin{equation} \label{eq:monomark}
 \mathbf{\bm\alpha}=(\delta_1,\delta_2,\cdots,\delta_d)^T.
\end{equation}
$\mathbf{\bm\alpha}$ is unique for one given ranking task and varies from task to task.
For a given ranking task with defined $\mathbf{\bm\alpha}$,
$\mathbf{x}$ precedes $\mathbf{y}$ for $x_j<y_j(j\in\mathbf{E})$ and $x_j>y_j(j\in\mathbf{F})$.

As $\mathbf{R}$ is totally ordered, we prefer to grade each point with a real value
to help with ranking.
Assume $\varphi:\mathbf{R}^d\mapsto\mathbf{R}$ is the ranking function to assign $\mathbf{x}$
a score which provides the ordering of $\mathbf{x}$.
$\varphi$ is required to be \textit{order-preserving} so that
$\varphi(\mathbf{x})$ has the same ordering in $\mathbf{R}$ as $\mathbf{x}$ in $\mathbf{R}^d$.
In order theory, an order-preserving function is also called \textit{isotone}
or \textit{monotone} \cite{priestley2002}.

\begin{definition}[\hspace{-0.01em}\cite{priestley2002}]\label{def:ordkeep}
 A function $\varphi:\mathbf{R}^d\mapsto\mathbf{R}$ is called
monotone (or, alternatively, order-preserving) if
\begin{equation}
 \mathbf{x}\preceq\mathbf{y} \Longrightarrow
\varphi(\mathbf{x})\leq\varphi(\mathbf{y})
\end{equation}
and strictly monotone if
\begin{equation} \label{eq:isotone}
 \mathbf{x}\preceq\mathbf{y}, \mathbf{x}\neq\mathbf{y} \Longrightarrow
\varphi(\mathbf{x})<\varphi(\mathbf{y})
\end{equation}
\end{definition}

Order-preserving is the basic requirement for a ranking function.
For a partially ordered set, $\varphi$ should assign a score to $\mathbf{x}$
no more than the score to $\mathbf{y}$ if $\mathbf{x}\preceq\mathbf{y}$.
Moreover, if $\mathbf{x}\neq\mathbf{y}$ also holds, the score assigned to $\mathbf{x}$
must be smaller than the score to $\mathbf{y}$.
As $\mathscr{A}$ is totally ordered and different points should be assigned with different scores,
the ranking function is required to be strictly monotone as stated by Eq.(\ref{eq:isotone}).
Otherwise, the ranking rule would be meaningless due to breaking the ordering
in original data space $\mathbf{R}^d$.

\begin{example}\label{emp:isotone}
 \emph{In addition to the two indicators in Example \ref{emp:gdp2leb}, another two indicators
 are taken to evaluate life qualities of countries:
 IMR\footnote{Infant Mortality Rate per 1000 born} and
 Tuberculosis\footnote{new cases of infectious Tuberculosis per 100,000 of population}.
 It is easily known that the life quality of one country would be higher if
 it has a higher LEB and GDP while a lower IMR and Tuberculosis.
 Let numerical observations on four countries to be
 $\mathbf{x}_I=(2.1, 62.7, 75, 59)$, $\mathbf{x}_M=(11.3, 75.5, 12, 30)$,
 $\mathbf{x}_G=(32.1, 79.2,6,4)$, and $\mathbf{x}_N=(47.6, 80.1, 3, 3)$ respectively.
 By Eq.(\ref{eq:orderDef}), they have the ordering
 $\mathbf{x}_I\preceq\mathbf{x}_M\preceq\mathbf{x}_G\preceq\mathbf{x}_N$ with
 ${\bm\alpha}=(1, 1, -1, -1)^T$.
 In this case, $\mathbf{E}=\{1,2\}$ and $\mathbf{F}=\{3,4\}$.
 Let $\varphi(\mathbf{x}_I)=0.407$, $\varphi(\mathbf{x}_M)=0.593$, $\varphi(\mathbf{x}_G)=0.785$
 and $\varphi(\mathbf{x}_N)=0.891$.
 Then $\varphi$ is a strictly monotone mapping which strictly preserves the ordering in $\mathbf{R}^4$.
 }
\end{example}

Recall that a differentiable function $f:\mathbf{R}\mapsto\mathbf{R}$ is nondecreasing if and
only if $f'(x)\geq0$ for all $x\in\mathbf{dom}f$, and increasing if $f'(x)>0$
for all $x\in\mathbf{dom}f$ (but the converse is not true) \cite{fitzpatrick2006}.
They are readily extended to the case of monotonicity in Definition \ref{def:ordkeep}
with respect to the order defined by Eq.(\ref{eq:orderDef}).

\begin{theorem}[\hspace{-0.01em}\cite{boyd2004}]\label{thm:monovecsuff}
  Let $\varphi:\mathbf{R}^d\mapsto\mathbf{R}$ be differentiable.
$\varphi$ is monotone if and only if
\begin{equation}
\nabla\varphi(\mathbf{x})\succeq\mathbf{0}
\end{equation}
where $\mathbf{0}$ is the zero vector.
$\varphi$ is strictly monotone if
\begin{equation}\label{eq:monocond1}
\nabla\varphi(\mathbf{x})\succ\mathbf{0}
\end{equation}
\end{theorem}

Theorem \ref{thm:monovecsuff} provides first-order conditions for monotonicity.
Note that `$\succ$' denotes a strict partial order \cite{boyd2004}.
Let
\begin{equation}
 \nabla\varphi(\mathbf{x})=\left(\frac{\partial\varphi}{\partial x_1},
 \frac{\partial\varphi}{\partial x_2},\cdots,\frac{\partial\varphi}{\partial x_d}\right)^T.
\end{equation}
$\nabla\varphi(\mathbf{x})\succ\mathbf{0}$ infers
$\frac{\partial\varphi}{\partial x_j}>0$ for $j\in\mathbf{E}$ and
$\frac{\partial\varphi}{\partial x_j}<0$ for $j\in\mathbf{F}$.
$\nabla\varphi(\mathbf{x})\succ\mathbf{0}$ infers that each
component of $\nabla\varphi(\mathbf{x})$ does not equal to zero. By
the case of strict monotonicity in Theorem \ref{thm:monovecsuff},
$\nabla\varphi(\mathbf{x})\succ\mathbf{0}$ infers not only that
$\varphi$ is strictly monotone from $\mathbf{R}^d$ to $\mathbf{R}$,
but also that the value $s=\varphi(\mathbf{x})$ is increasing with
respect to $x_j(j\in\mathbf{E})$ and decreasing with respect to
$x_j(j\in\mathbf{F})$. Vice versa, if
$\frac{\partial\varphi}{\partial x_j}$ is bigger than zero for
$j\in\mathbf{E}$ and smaller than zero for $j\in\mathbf{F}$,
$\nabla\varphi(\mathbf{x})\succ\mathbf{0}$ holds and infers
$\varphi$ is a strictly monotone mapping. Lemma \ref{lem:mono} can
be concluded immediately.

\begin{lemma}\label{lem:mono}
 $s=\varphi(\mathbf{x})$ is strictly monotone if and only if $s$ is strictly monotone along
 $x_i$ with fixed the others $x_j(j\neq i)$.
\end{lemma}

Further more, a strictly monotone mapping infers a \textit{one-to-one} mapping that for
a value $s\in\mathbf{rang}\varphi$ there is exactly one point
$\mathbf{x}\in\mathbf{dom}\varphi$ such that $\varphi(\mathbf{\mathbf{x}})=s$.
If the point $\mathbf{x}$ is denoted by
$\mathbf{x}=\mathbf{f}(s)$, $\mathbf{f}:\mathbf{R}\mapsto\mathbf{R}^d$ is called the
\textit{inverse} mapping of $\varphi$ and inherits the property of strict monotonicity
of its origin $\varphi$.

\begin{theorem}\label{thm:monoinverse}
 Assume $\nabla\varphi(\mathbf{x})\succ\mathbf{0}$.
 There exists an inverse mapping denoted by
 $\mathbf{f}:\mathbf{rang}\varphi\mapsto\mathbf{dom}\varphi$ such that
 $\nabla\mathbf{f}(s)\succ\mathbf{0}$ holds for all $s\in\mathbf{rang}\varphi$,
 that is for $\forall s_1,s_2\in\mathbf{rang}\varphi$
 \begin{equation}
  s_1<s_2 \Longrightarrow \mathbf{f}(s_1)\preceq\mathbf{f}(s_2), \mathbf{f}(s_1)\neq\mathbf{f}(s_2).
 \end{equation}
\end{theorem}

Proof of Theorem \ref{thm:monoinverse} can be found in Appendix \ref{app:monoproof}.
The theorem also holds in the other direction.
Assuming $\mathbf{f}:\mathbf{R}\mapsto\mathbf{R}^d$, if $\nabla\mathbf{f}(s)\succ\mathbf{0}$,
there exists an inverse mapping $\varphi:\mathbf{rang}\mathbf{f}\mapsto\mathbf{dom}\mathbf{f}$ and
$\nabla\varphi(\mathbf{x})\succ\mathbf{0}$ holds for all $\mathbf{x}\in\mathbf{rang}\mathbf{f}$.
Because of the one-to-one correspondence, $\mathbf{f}$ and $\varphi$ share the same
geometric properties such as scale and translation invariance, smoothness and strict monotonicity
\cite{fitzpatrick2006}.

\section{Meta-Rules} \label{sec:metarule}

As a ranking function for $\varphi:\mathbf{R}^d\mapsto\mathbf{R}$,
$\varphi(\mathbf{x})$ outputs a real value $s=\varphi(\mathbf{x})$ as the ranking score
for a given point $\mathbf{x}$.
The ranking list of objects would be provided by sorting their ranking scores
in ascending/descending order.
Since unsupervised ranking has no label information to verify the ranking list,
we restrict ranking functions with five essential features to guarantee
that a reasonable ranking list is provided.
These features are capable of serving as high-level guidance of modeling ranking functions.
They are also capable of serving as high-level assessments for unsupervised ranking performance,
different from assessments for supervised ranking performance which take qualities of ranking labels.
Any functions from $\mathbf{R}^d$ to $\mathbf{R}$ with all the five features can serve as
ranking functions and be able to provide a reasonable ranking list.
These features are rules for ranking rules, namely \textit{meta-rules}.

\subsection{Scale and Translation Invariance}

 \begin{definition}[\hspace{-0.01em}\cite{cambini2003}]
  A ranking rule is invariant to scale and translation if for $\mathbf{x}\preceq\mathbf{y}$
  \begin{equation}
    \varphi(\mathbf{x})\leq\varphi(\mathbf{y})
    \Longleftrightarrow\varphi(\mathbf{L}(\mathbf{x}))
    \leq\varphi(\mathbf{L}(\mathbf{y})).
  \end{equation}
 where $\mathbf{L}(\cdot)$ performs scale and translation.
 \end{definition}

 Numerical observations on different indicators are taken on different dimensions of quantity.
 In Example \ref{emp:gdp2leb}, GDP is measured in thousands of dollars
 while LEB ranges from 40 to 90 years. They are not in the same dimensions of quantity.
 As a general data preprocessing technique,
 scale and translation can take them into the same dimensions (e.g. $[0,1]$)
 while preserving their original ordering.
 If let $\mathbf{L}$ be a linear transformation on $\mathbf{R}^d$, we have
 $\mathbf{x}\preceq\mathbf{y}\Longleftrightarrow\mathbf{L}(\mathbf{x})\preceq\mathbf{L}(\mathbf{y})$
 for $\mathbf{x},\mathbf{y}\in\mathbf{R}^d$ \cite{cambini2003}.
 Therefore, a ranking function $\varphi(\mathbf{x})$ should produce the same ranking list
 before and after scaling and translating.

\subsection{Strict Monotonicity}

 \begin{definition}[\hspace{-0.01em}\cite{priestley2002}]\label{def:monofeature}
   $\varphi(\mathbf{x})$ is strictly monotone if
  $\varphi(\mathbf{x}_i)<\varphi(\mathbf{x}_j)$
  for $\mathbf{x}_i\preceq\mathbf{x}_j$ and $\mathbf{x}_i\neq\mathbf{x}_j(i\neq{j})$ .
 \end{definition}

 Strict monotonicity in Definition \ref{def:ordkeep} is specified here as one of meta-rules
 for ranking.
 For ordinal classification problem, monotonicity is a general constraint since two different objects
 would be classified into the same class \cite{kotlow2013}.
 But for the ranking problem discussed in this paper, it requires the strict monotonicity
 since different objects should have different scores for ranking.
 $\varphi(\mathbf{x}_i)=\varphi(\mathbf{x}_j)$ holds
 if and only if $\mathbf{x}_i=\mathbf{x}_j(i\neq{j})$.
 In Example \ref{emp:gdp2leb}, $\mathbf{x}_1\preceq\mathbf{x}_2$ and
 $\mathbf{x}_i\neq\mathbf{x}_j$ indicate that a higher score should be assigned to $\mathbf{x}_2$
 than $\mathbf{x}_1$. And so do $\mathbf{x}_3$ and $\mathbf{x}_4$.
 Therefore, the ranking function $\varphi(\mathbf{x})$ is required to be a strictly monotone mapping.
 Otherwise, the ranking list would be not convincing.
 $\varphi$ in Example \ref{emp:isotone} is to the point referred here.

\subsection{Linear/Nonlinear Capacities}

 \begin{definition}
  $\varphi(\mathbf{x})$ has the capacities of linearity and nonlinearity
  if $\varphi(\mathbf{x})$ is able to depict the relationship of both linearity and nonlinearity.
 \end{definition}

 Taking the ranking task in Example \ref{emp:gdp2leb} for illustration,
 one has no knowledge about the relationship between LEB and the score.
 The score might be a either linear or nonlinear function of LEB.
 It is the similar case for the relationship between GDP and the score.
 Therefore, $t=\varphi(\mathbf{x})$ should embody both of the linear
 and nonlinear relationships between $t$ and $x_j$.
 For the ranking task in Example \ref{emp:gdp2leb},
 the ranking function $\varphi$ should be a linear function of LEB for fixed GDP
 if LEB is linear with $t$.
 Meanwhile, $\varphi$ should also be a nonlinear function of GDP for fixed LEB
 if GDP is nonlinear with $t$.

\subsection{Smoothness}

 \begin{definition}[\cite{fitzpatrick2006}]
  $\varphi(\mathbf{x})$ is smooth if $\varphi(\mathbf{x})$ is $\mathscr{C}^h(h\geq1)$.
 \end{definition}

 In mathematical analysis, a function is called smooth if it has
 derivatives of all orders \cite{fitzpatrick2006}.
 Yet a ranking function $\varphi(\mathbf{x})$ is required to be
 of class $\mathscr{C}^h$ where $h\geq1$.
 That is, $\varphi(\mathbf{x})$ is continuous and has the first-order derivative
 $\nabla\varphi(\mathbf{x})$. The first-order derivative
 $\nabla\varphi(\mathbf{x})$ guarantees that $\varphi(\mathbf{x})$ will exert
 a consistent ranking rule for all objects and the ranking rule
 would be not abruptly changed for some object.
 Taking the polyline in Fig. \ref{fig:drawback} for illustration,
 it is of class $\mathscr{C}^0$ but not of class $\mathscr{C}^1$ because it
 is continuous but not differentiable at the connecting vertex of the
 two lines. This would lead to an unreasonable ranking for those
 points projected to the vertex.

\subsection{Explicitness of Parameter Size}

  \begin{definition}
   $\varphi(\mathbf{x})$ has the property of explicitness if
   $\varphi(\mathbf{x})$ has known parameter size for a fair comparison among ranking models.
  \end{definition}

 Hu et al. \cite{hubg2009} considered that nonparametric approaches are a class of ``black-box''
 approaches since they can not be interpreted by our intuition.
 As a ranking function, $\varphi(\mathbf{x})$ should be semantically interpretable
 so that $\varphi(\mathbf{x})$ has systematical meanings.
 For example, $\varphi(\mathbf{x})={\bm\theta}^T\mathbf{x}$ gives explicitly
 the linear expression with parameter size $d$ which is the dimension of the parameter $\bm\theta$.
 It can be interpreted that the score of $\mathbf{x}$ is linear with $\mathbf{x}$ and the parameter
 $\bm\theta$ is the allocation proportion vector of indicators for ranking.
 Moreover, if there is another ranking model with the same characteristics,
 $\varphi(\mathbf{x},{\bm\theta})$ would be more applicable if it has a smaller size of parameters.

These five meta-rules above is the guidance of designing a
reasonable and practical ranking function. To perform a ranking
task, a ranking function should satisfy all the five meta-rules
above to produce a convincing ranking list. Any ranking function
that breaks any of them would produce a biased and unreasonable
ranking list. In this sense, they can be regarded as high-level
assessments for unsupervised ranking performance.

\section{Ranking Principal Curves} \label{sec:rpc}

In this section, we propose a \textit{ranking principal curve} (RPC) model
to perform an unsupervised ranking task with a principal curve which has all the five meta-rules.
The RPC is parametrically designed to a cubic B\'{e}zier curve with control points restricted
in the interior of a hypercube.

\subsection{RPC Model}

\begin{figure}[t]
\centering
\includegraphics[scale=0.5, bb = 0 0 430 380]{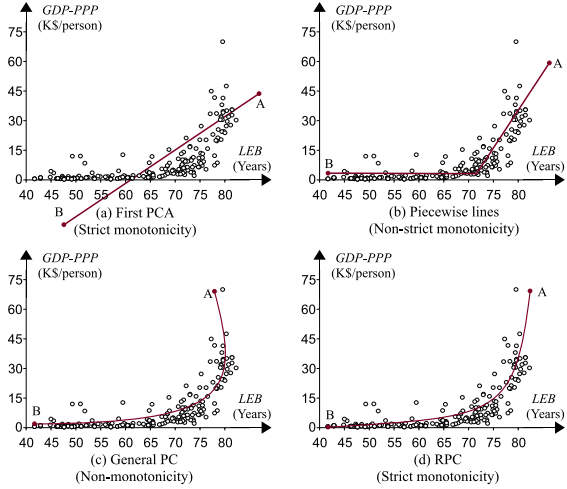}
\caption{Schematic plots of ranking skeletons (heavy solid lines or
curves in red). Circle points: observations of countries on two
indicators: LEB and GDP. } \label{fig:metarule}
\end{figure}

The simplest ranking rule is the first PCA which summarizes the data
in $d$-dimensional space with the largest principal component line
\cite{anderson2003}. The first PCA seeks the direction $\mathbf{w}$
that explains the maximal variance of the data cloud. Then
$\mathbf{x}$ is orthogonally projected by $\mathbf{w}^T\mathbf{x}$
onto the line passing through the mean ${\bm\mu}$. The line can be
regarded as the \textit{ranking skeleton}. Projected points take an
ordering along the ranking skeleton which is just the ordering of
their first principal components computed by
$\mathbf{w}^T\mathbf{x}$. Let $s=\mathbf{w}^T\mathbf{x}$ and an
ordering of $s$ gives the ordering of $\mathbf{x}$. As a ranking
function, the first PCA is smooth, explicitly expressed, and
invariant to scale and translation. It works well for the skeleton
of slender ellipse distributing data. However, the first PCA can
hardly depict the skeleton of data distributions like crescents
(Fig. \ref{fig:metarule}(a)) such that the produced ranking list is
not convincing. What's more, the first PCA might be non-strictly
monotone when the direction $\mathbf{w}$ is parallel to one
coordinate axis such that it can not discriminate those points like
$\mathbf{x}_1$ and $\mathbf{x}_2$ in Example \ref{emp:gdp2leb} since
they will be projected to the same points if the first PCA is on the
direction parallel to the horizontal line. The problems referred
above hinder the first PCA from extensive applications in
comprehensive evaluation.

Recalling that principal curves are nonlinear extensions of the first PCA \cite{hastie1989},
we try to summarize multiple indicators in the data space with a principal curve
(Appendix \ref{app:pc} gives a brief review of principal curves).
Assuming $\mathbf{f}(s, {\bm\theta})(\bm{\theta}\in\mathbf{\Theta})$ is the principal curve
of a given data cloud, it provides an ordering of projected points on the principal curve,
in a way similar to the first PCA.
Intuitively, the principal curve is a good choice to perform ranking tasks.
On the one hand, unsupervised ranking could only rely on those numerical observations for
ranking candidates on given attributes.
For the dataset with a linking network, PageRank can calculate a score with backlinks
for each point \cite{page1998}.
When there is no link between points, a score can still be calculated according to
the ranking skeleton, instead of link structure.
On the other hand, the principal curve reconstructs $\mathbf{x}$ according to
$\mathbf{x}=\mathbf{f}(s,{\bm\theta})+{\bm\varepsilon}$, instead of
$\mathbf{x}=\mathbf{\bm\mu}+s\mathbf{w}+{\bm\varepsilon}$ for the first PCA.
To perform ranking tasks, a ranking function assigns a score $s$ to $\mathbf{x}$ by
$s=\varphi(\mathbf{x},{\bm\theta})$. Actually, noise is inevitable due to measuring errors and
influence from exclusive indicators from $\mathbf{V}$.
Thus the latent score should be produce after removing noise from $\mathbf{x}$, that is
$s=\varphi(\mathbf{x}-{\bm\varepsilon},{\bm\theta})$.
As a ranking function, $\varphi$ is assumed to be strictly monotone.
Thus, data points and scores are one-to-one correspondence
and there exists an inverse function $\mathbf{f}$ for $\varphi$ such that
\begin{equation}\label{eq:pc&noise}
  \mathbf{x}=\mathbf{f}(s,{\bm\theta})+{\bm\varepsilon}
\end{equation}
which is the very principal curve model\cite{hastie1989}. The
inverse function can be taken as the generating function for
numerical observations from the score $s$ which can be regarded to
be pre-existing.

As stated in Section \ref{sec:metarule}, there are five meta-rules
for a function $\varphi(\mathbf{x},{\bm\theta})$ to serve as a
ranking rule. As $\varphi(\mathbf{x},{\bm\theta})$ is required to be
strictly monotone, there exists an inverse function
$\mathbf{f}(s,{\bm\theta})$ which is also strictly monotone by
Theorem \ref{thm:monoinverse}. Correspondingly, $\varphi$ and its
inverse $\mathbf{f}$ share the other properties of scale and
translation invariance, smoothness, capacities of linearity and
nonlinearity, and explicitness of parameter size. A principal curve
should also follow all the five meta-rules to serve as a ranking
function. However, polyline approximations of the principal curve
might go against smoothness and strict monotonicity (e.g. Fig.
\ref{fig:metarule}(b)). A smooth principal curve would also go
against strict monotonicity (e.g. Fig. \ref{fig:metarule}(c)). Both
of them would make unreasonable ranking solutions as illustrated in
Example \ref{emp:gdp2leb}. Within the framework of Fig.
\ref{fig:rpcmodel}, all the five meta-rules can be modeled as
constraints to the ranking function. Since a principal curve is
defined to be smooth and invariant to scale and translation
\cite{hastie1989}, the constraint of strict monotonicity would make
it be capable of performing ranking tasks (e.g. Fig.
\ref{fig:metarule}(d)). Naturally, the principal curve should have a
known parameter size for interpretability reason. We present
Definition \ref{def:rpc} for unsupervised ranking with a principal
curve.

\begin{definition} \label{def:rpc}
 A curve $\mathbf{f}(\mathbf{x},{\bm\theta})$ in $d$-dimensional space is called a
\textit{ranking principal curve} (RPC)
if $\mathbf{f}(\mathbf{x},{\bm\theta})$ is a strictly monotone principal curve of given data cloud
and it is explicitly expressed with known parameters $\bm\theta$ of limited size.
\end{definition}

\subsection{RPC Formulation with B\'{e}zier Curves} \label{ssec:rpc2bc}

To perform a ranking task, a principal curve model should follow all
the five meta-rules (Section \ref{sec:metarule}) which can be also
similarly defined for $\mathbf{f}$. However, not all of principal
curve models can perform ranking tasks. The models in
\cite{hastie1989,brpc1992,delicado2001,einbeck2005} lack of
explicitness and can not make a monotone mapping on $\mathbf{R}^d$
(Fig. \ref{fig:metarule}(c)). Polyline approximation
\cite{kegl2000,chang2001,gorban2010} misses the requirements for
smoothness and strictly monotonicity (Fig. \ref{fig:metarule}(b)). A
new principal curve model is needed to perform ranking while
following all the five meta-rules.

In this paper, an RPC is parametrically modeled with
a B\'{e}zier curve
\begin{equation} \label{eq:bc}
 \mathbf{f}(s) = \sum_{r=0}^k B_r^k(s) \mathbf{p}_r,s\in[0,1]
\end{equation}
which is formulated in terms of Bernstein polynomials \cite{farin1997}
\begin{eqnarray}
B_r^k(s) &=& \binom{k}{r}(1-s)^{k-r}s^r, \label{eq:bernstein} \\
\binom{k}{r} &=& \frac{k!}{r!(k-r)!}. \label{eq:binom}
\end{eqnarray}
In Eq.(\ref{eq:bc}), $\mathbf{p}_r\in{\mathbf{R}^d}$ are control and
end points of the B\'{e}zier curve which are in the place of the
function parameters $\bm\theta$ in Eq.(\ref{eq:pc&noise}).
Particularly, when $k=3$, Eq.(\ref{eq:bc}) has the matrix form of
\begin{equation}\label{eq:bcmatrixform}
 \mathbf{f}(s)=\mathbf{PMz}.
\end{equation}
where
\begin{eqnarray*}
\mathbf{P}\hspace{-0.25cm}&=& \hspace{-0.25cm}(\mathbf{p}_0,\mathbf{p}_1,\mathbf{p}_2,\mathbf{p}_3) \\
 \mathbf{M}\hspace{-0.25cm}&=& \hspace{-0.25cm}\left(\begin{array}{cccc}
            1 & -3 & 3 & -1 \\
        0 & 3 & -6 & 3 \\
        0 & 0 & 3 & -3 \\
        0 & 0 & 0 & 1
           \end{array}\right),
\mathbf{z}=\left(\begin{array}{c}
            1\\ s\\ s^2\\ s^3
           \end{array}\right)
\end{eqnarray*}
In case $k>3$, the model would become more complex and bring about overfitting problem.
In case $k<3$, the model is too simple to represent all possible monotonic curves.
$k=3$ is the most suitable degree to perform the ranking task.

A cubic B\'{e}zier curve with constraints on control points can be
proved to have all the five meta-rules. First of all, the
formulation Eq.(\ref{eq:bc}) is a nonlinear interpolation of control
points and end points in terms of Bernstein polynomials
\cite{farin1997}. These points are the determinant parameters of
total size $d\times4$. Different locations of these points would
produce different shapes of nonlinear curves besides straight lines
\cite{hubg1998}. Scale and translation to B\'{e}zier curves are
applied to these points without changing the ranking score which is
contained in $\mathbf{z}$
\begin{equation} \label{eq:invariance}
 \mathbf{\Lambda}\mathbf{f}(s)+{\bm\beta}
 =\mathbf{\Lambda}\mathbf{PMz}+{\bm\beta}
 =(\mathbf{\Lambda}\mathbf{P}+{\bm\beta})\mathbf{Mz}
\end{equation}
where $\mathbf{\Lambda}$ is a diagonal matrix with scaling factors to dimensions
and ${\bm\beta}$ is the translation vector.
This property allows us to put all data into $[0,1]^d$ in order to facilitate ranking.
What's more, the derivative of $\mathbf{f}(s)$ is a lower order B\'{e}zier curve
\begin{equation}\label{eq:bcderivative}
 \frac{d\mathbf{f}(s)}{ds} = k\sum_{j=0}^{k-1}B_j^{k-1}(s)(\mathbf{p}_{j+1}-\mathbf{p}_j)
\end{equation}
which involves the calculation of end points and control points.
Its derivatives of all orders exist for all $s\in[0,1]$ and thus Eq.(\ref{eq:bc}) is smooth enough.
Last but not the least, it has been proved that a cubic B\'{e}zier curve can
perform the four basic types of strict monotonicity in two-dimensional space \cite{hubg1998}.
Let end points after scale and translation
are denoted by $\mathbf{p}_0=\frac{1}{2}(\mathbf{1}-\mathbf{\bm\alpha})$
and $\mathbf{p}_3=\frac{1}{2}(\mathbf{1}+\mathbf{\bm\alpha})$.
Control points $\mathbf{p}_2$ and $\mathbf{p}_3$ are the determinants for nonlinearity of the
cubic B\'{e}zier curve (Fig. \ref{fig:bcshape}).
In two-dimensional space, $\mathbf{f}(s)$ is proved to be increasing along each coordinate
if control points are restricted in the interior of the hypercube $[0,1]^d$ \cite{hubg1998}.
Thus, a proposition can be deduced by Lemma \ref{lem:mono}.

\begin{proposition} \label{prop:mono}
 $\mathbf{f}(s)$ is strictly monotone for $s\in[0,1]$ with
 $\mathbf{p}_0=\frac{1}{2}(\mathbf{1}-\bm\alpha)$,
 $\mathbf{p}_3=\frac{1}{2}(\mathbf{1}+\bm\alpha)$ and
 $\mathbf{p}_1,\mathbf{p}_2\in(0,1)^d$.
\end{proposition}

What is the most important, there always exists an RPC parameterized by
a cubic B\'{e}zier curve which is strictly monotone for a group of numerical observations.
The existence has failed to be proved in many principal curve models
\cite{hastie1989,chang2001,gorban2010}.

\begin{theorem}\label{thm:existence}
 Assume that $\mathbf{x}$ is the numerical observation of a ranking candidate
 and that $E\|{\mathbf{x}}\|^2<\infty$.
 There exists $\mathbf{P}^*\in[0,1]^d$ such that
 $\mathbf{f}^*(s)=\mathbf{P}^*\mathbf{Mz}$ is strictly monotone and
\begin{equation}\label{eq:exist}
 J(\mathbf{P}^*) = \inf\left\{J(\mathbf{P})
= E\left(\inf_s\|{\mathbf{x}-\mathbf{PMz}}\|^2\right)\right\}.
\end{equation}
\end{theorem}

Proof of Theorem \ref{thm:existence} can be found in Appendix \ref{app:existproof}.

\section{RPC Learning Algorithm} \label{sec:rpclearning}

To perform unsupervised ranking from the numerical observations of
ranking candidates
$\mathbf{X}=(\mathbf{x}_1,\mathbf{x}_2,\cdots,\mathbf{x}_n)$, we
should first learn control points of the curve in Eq.(\ref{eq:bc}).
The optimal points achieve the infimum of the estimation of
$J(\mathbf{P})$ in Eq.(\ref{eq:exist}). By the principal curve
definition proposed by Hastie et al.\cite{hastie1989}, the RPC is
the curve which minimizes the summed residual ${\bm\varepsilon}$.
Therefore, the ranking task is formulated as a nonlinear
optimization problem
\begin{eqnarray}
\min
& &J(\mathbf{P},\mathbf{s})=\sum_{i=1}^n \| \mathbf{x}_i - \mathbf{PMz}_i \|^2
\label{eq:msebc}\\
s.t.& & \left.\left(\frac{\partial\mathbf{PMz}}{\partial s}\right)^T
\left(\mathbf{x}_i - \mathbf{PMz}\right)\right|_{s=s_i}=0, \label{eq:orth} \\
& & \mathbf{s}=(s_1,s_2,\cdots,s_n) ,\mathbf{z}=(1,s,s^2,s^3)^T \nonumber \\
& & \mathbf{P}\in{[0,1]^{d\times 4}}, \quad s_i\in[0,1], \nonumber \\
& & i=1,2,\cdots,n \nonumber
\end{eqnarray}
where Eq.(\ref{eq:orth}) determines $s_i$ to find the point on the curve which has
the minimum residual to reconstruct $\mathbf{x}_i$ by $\mathbf{f}(s_i)$.
Obviously, a local minimizer $(\mathbf{P}^*,\mathbf{s}^*)$ can be achieved in an
alternating minimization way
\begin{eqnarray}
\hspace{-1cm}& & \mathbf{P}^{(t+1)} = \arg\min \sum_{i=1}^n \| \mathbf{x}_i - \mathbf{PMz}_i^{(t)}
\|^2 \label{eq:updateP}\\
\hspace{-1cm}& & \left.\left(\frac{\partial \mathbf{P}^{(t+1)}\mathbf{Mz}}
{\partial s}\right)^T\left(\mathbf{x}_i -
\mathbf{P}^{(t+1)}\mathbf{Mz}\right)
\right|_{s=s_i^{(t+1)}}=0 \label{eq:updateT}
\end{eqnarray}
where $t$ means the $t$th iteration.

The optimal solution of Eq.(\ref{eq:updateP}) has an explicit expression.
Associate $\mathbf{X}$ with $\mathbf{Z}$
\begin{equation}
 \mathbf{Z} = \left(\begin{array}{cccc}
       1 & 1 & \cdots & 1\\
       s_1 & s_2 & \cdots & s_n\\
       s_1^2 & s_2^2 & \cdots & s_n^2\\
       s_1^3 & s_2^3 & \cdots & s_n^3
      \end{array}\right) =
\left(\mathbf{z}_1, \mathbf{z}_2, \cdots, \mathbf{z}_n\right)
\end{equation}
and Eq.(\ref{eq:msebc}) can be rewritten in matrix form
\begin{eqnarray}
 J(\mathbf{P}, \mathbf{s})
&=& \| \mathbf{X}-\mathbf{PMZ} \|_F \nonumber \\
&=&tr(\mathbf{X}^T\mathbf{X})-2tr(\mathbf{PMZ}\mathbf{X}^T) \nonumber \\
& & + tr(\mathbf{PMZZ}^T\mathbf{M}^T\mathbf{P}^T). \label{eq:msematrix}
\end{eqnarray}
Setting the derivative of $J$ with respect to $\mathbf{P}$ to zero
\begin{equation}
 \frac{\partial J}{\partial \mathbf{P}}=
2\left(\mathbf{P}(\mathbf{MZ})(\mathbf{MZ})^T-\mathbf{X}(\mathbf{MZ})^T\right)=0
\end{equation}
and remembering $A^+=A^T(AA^T)^+$ \cite{roger1985}, we get an explicitly expression
for the minimum point of Eq.(\ref{eq:msebc})
\begin{equation} \label{eq:pinv}
 \mathbf{P}=\mathbf{X}(\mathbf{MZ})^T\left((\mathbf{MZ})(\mathbf{MZ})^T\right)^+
=\mathbf{X}(\mathbf{MZ})^+
\end{equation}
where $(\cdotp)^+$ takes pseudo-inverse computation. Based on the
$t$th iterative results $\mathbf{Z}^{(t)}$, the optimal solution can
be given by substituting $\mathbf{Z}^{(t)}$ into Eq.(\ref{eq:pinv})
which is $\mathbf{P}^{(t+1)}=\mathbf{X}(\mathbf{MZ}^{(t)})^+$.
However, $(\mathbf{MZ}^{(t)})^+$ is computationally expensive in
numerical experiments and $\mathbf{X}$ is always ill-conditioned
which has a high condition number, resulting in that a very small
change in $\mathbf{Z}^{(t)}$ would produce a tremendous change in
$\mathbf{P}^{(t+1)}$. $\mathbf{Z}^{(t)}$ is not the optimal solution
of Eq.(\ref{eq:msebc}) but a intermediate result of the iteration,
and $\mathbf{P}^{(t+1)}$ would thereby go far away from the optimal
solution. To settle out the problem, we employ the Richardson
iteration \cite{Richard1910} with a preconditioner $\mathbf{D}$
which is a diagonal matrix with the $L_2$ norm of columns of
$(\mathbf{MZ}^{(t)})(\mathbf{MZ}^{(t)})^T$ as its diagonal elements.
Then $\mathbf{P}^{(t+1)}$ is updated according to
\begin{eqnarray}
 \mathbf{P}^{(t+1)}
 &=& \mathbf{P}^{(t)}-\gamma^{(t)}(\mathbf{P}^{(t)}(\mathbf{MZ}^{(t)})
(\mathbf{MZ}^{(t)})^T \nonumber \\
 & & -\mathbf{X}(\mathbf{MZ}^{(t)})^T)\mathbf{D}^{-1} \label{eq:numiterP}
\end{eqnarray}
where
$\gamma^{(t)}$ is a scalar parameter such that the sequence $\mathbf{P}^{(t)}$ converges.
In practice, we set
\begin{equation}
\gamma^{(t)} = \frac{2}{\lambda_{min}^{(t)}+\lambda_{max}^{(t)}}
\end{equation}
where $\lambda_{min}^{(t)}$ and $\lambda_{max}^{(t)}$ is the minimum and maximum
eigenvalues of $(\mathbf{MZ}^{(t)})(\mathbf{MZ}^{(t)})^T$ respectively \cite{golub1996}.

After getting $\mathbf{P}^{(t+1)}$, the score vector $\mathbf{s}^{(t+1)}$ can be calculated as
the solution to Eq.(\ref{eq:updateT}).
Eq.(\ref{eq:updateT}) is a quintic polynomial equation which rarely has explicitly expressed roots.
In \cite{pastva1998}, $s_i$ for $\mathbf{x}_i$ was approximated by Gradient and
Gauss-Newton methods respectively. Jenkins-Traub method \cite{jenkins1970} was also considered
to find the roots of the polynomial equation directly.
As Eq.(\ref{eq:orth}) is designed to find the minimum distance of point $\mathbf{x}_i$ from the
curve, we adopt Golden Section Search (GSS) \cite{bazaraa2006}
to find the local approximate solution to Eq.(\ref{eq:updateT}).

\begin{algorithm}[t]
\caption{Algorithm to learn an RPC.}
\label{agrm:RPC}
\begin{algorithmic}[1]
\REQUIRE  ~~
\\
$\mathbf{X}$: data matrix; \\
$\xi$: a small positive value;
\ENSURE ~~ \\
$\mathbf{P}^*$: control points of the learned B\'{e}zier curve \\
$\mathbf{s}^*$: the score vector of objects in the set. \\
\STATE Normalize $\mathbf{X}$ into $[0,1]^d$;
\STATE Initialize $\mathbf{P}^{(0)}$;  \label{step:ini}\\
\WHILE {$\vartriangle{J}>\xi$}
  \STATE Adopt GSS to find the approximate solution $\mathbf{s}^{(t)}$; \\
  \STATE Compute $\mathbf{P}^{(t+1)}$ using a preconditioner;\\
  \IF {$\vartriangle{J}<0$} \label{step:break}
      \STATE break;
  \ENDIF
\ENDWHILE
\end{algorithmic}
\end{algorithm}

Algorithm \ref{agrm:RPC} summarizes the alternative optimization procedure.
Before performing the ranking task, numerical observations of objects should be normalized
into $[0,1]^d$ by
\begin{equation} \label{eq:normalization}
  \hat{\mathbf{x}}=\frac{\mathbf{x}-\mathbf{x}_{min}}{\mathbf{x}_{max}-\mathbf{x}_{min}}
\end{equation}
where $\hat{\mathbf{x}}$ is the normalized vector of $\mathbf{x}$,
$\mathbf{x}_{min}$ the minimum vector and $\mathbf{x}_{max}$ the maximum vector.
Grading scores would be unchanged as scaling and translating are only performed on
control points and end points (Eq.(\ref{eq:invariance})) without changing the interpolation values.
In Step \ref{step:ini}, we initialize the end points
as $\mathbf{p}_0=\frac{1}{2}(\mathbf{1}-\mathbf{\bm\alpha})$
and $\mathbf{p}_3=\frac{1}{2}(\mathbf{1}+\mathbf{\bm\alpha})$,
and randomly select samples as control points.
During learning procedure, $\mathbf{P}^{(t)}$ is automatically learned
making a B\'{e}zier curve to be an RPC in numerical experiments.
In Step \ref{step:break}, $\vartriangle{J}<0$ occurs when
$J$ begins to increase. In this case, the algorithm
stops updating $(\mathbf{P}^{(t)},\mathbf{s}^{(t)})$ and gets a local minimum $J$.
Proposition \ref{prop:converge} guarantees the convergency of
the sequence found by RPC learning algorithm (proof can be found in
Appendix \ref{app:propproof}).
Therefore, the RPC learning algorithm finds a converging sequence of
$(\mathbf{P}^{(t)}, \mathbf{s}^{(t)})$ to achieve the infimum in Eq.(\ref{eq:exist}).

\begin{proposition}\label{prop:converge}
If $\mathbf{P}^{(t)}\rightarrow\mathbf{P}^*$ as $t\rightarrow\infty$,
$J(\mathbf{P}^{(t)},\mathbf{s}^{(t)})$ is a decaying sequence which converges to
$J(\mathbf{P}^*,\mathbf{s}^*)$ as $t\rightarrow\infty$.
\end{proposition}

Algorithm \ref{agrm:RPC} converges in limited steps.
In each step, $\mathbf{P}$ is updated in $4\times{d}$ size and scores for
points are calculated in $n$ size.
When iteration stops, ranking scores are produced along with $\mathbf{P}$.
In summary, the computational complexity of RPC unsupervised ranking model is $O(4d+n)$.
Compared to the ranking rule of weighted summation, ranking with RPC model costs a little more.
However, weighted summation needs weight assignments by a domain expert
such that it is more subjective because weights is diverse expert by expert.
But RPC model needs no expert to assign weight proportions to indicators.
The learning procedure of RPC model does the whole work for ranking.

The RPC learning algorithm learns a ranking function in a completely
different way from the traditional methods. On the one hand, the
ranking function is in constraints of five meta-rules for ranking
rules. Integrating meta-rules with ranking functions makes the
ranking rule be more in line with human knowledge about ranking
problems. As a high level knowledge, these meta-rules are capable of
evaluating ranking performance. On the other hand, ranking is
carried out following the principle of unsupervised ranking,
``\textit{let the data speak for themselves}''. For unsupervised
ranking, there is no information for ranking labels to guide the
system to learn a ranking function. As a matter of fact, the
structure of the dataset contains the ordinal information between
objects. If all the determining factors of ordinal relations are
included, the RPC can thread through all the objects successively.
In practice, the most influential indicators are selected to
estimate the order of objects, but the rest factors still affect the
numerical observation. In the case we know nothing about the rest
factors, we would better to minimize the effect which we formulate
to be error ${\bm\varepsilon}$. Therefore, minimizing errors is
adopted as the learning objection in case no ranking label can be
available.

\section{Experiments} \label{sec:applications}

\subsection{Comparisons with Ranking Aggregation} \label{ssec:comp2rankagg}

\begin{figure}[t]
\centering
\includegraphics[scale=0.25, bb = 45 50 547 546]{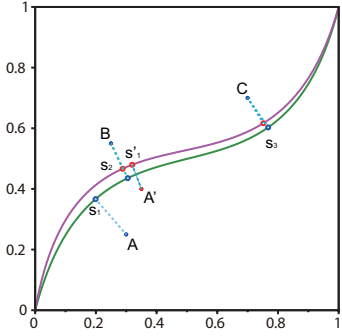}
\caption{$A$, $B$ and $C$ are three objects to rank.
$s_1$, $s_2$ and $s_3$ are scores given by the RPC (in green) of S-type shape in the figure.
A different observation of $A$ (denoted by $A'$) would give a different RPC (in pink)
and thus a different ordering of objects. }
\label{fig:illustrate}
\end{figure}

\begin{table*}[t] \scriptsize
 \centering
\caption{RPC model can detect ordinal information contained in numerical observations in
Fig. \ref{fig:illustrate}. }
\begin{threeparttable}
\vspace{-2em}
\subtable[A group of bservations and ranking lists by different rules]{
\begin{tabular}{|c||cc|cc|c|cc|}
\hline
\multirow{2}{*}{\textbf{Object}} &\multicolumn{2}{c|}{$x_1$}  & \multicolumn{2}{c|}{$x_2$}
& \multirow{2}{*}{\textbf{RankAgg}} & \multicolumn{2}{c|}{\textbf{RPC}}\\
\cline{2-5} \cline{7-8}
 & \textbf{Value} & \textbf{Order} & \textbf{Value} & \textbf{Order} &
 & \textbf{Score} & \textbf{Order} \\
\hline
$A$  & $0.3$ & 2 & $0.25$ & 1 & 1.5 & 0.2329 & 1\\
\hline
$B$  & $0.25$ & 1 & $0.55$ & 2 & 1.5 & 0.3304 & 2\\
\hline
$C$  & $0.7$ & 3 & $0.7$ & 3 & 3 & 0.7300 & 3\\
\hline
\end{tabular}
\label{tab:Cb}
}
\\
\subtable[Another group of bservations and ranking lists by different rules]{
\begin{tabular}{|c||cc|cc|c|cc|}
\hline
\multirow{2}{*}{\textbf{Object}} &\multicolumn{2}{c|}{$x_1$}  & \multicolumn{2}{c|}{$x_2$}
& \multirow{2}{*}{\textbf{RankAgg}} & \multicolumn{2}{c|}{\textbf{RPC}}\\
\cline{2-5} \cline{7-8}
 & \textbf{Value} & \textbf{Order} & \textbf{Value} & \textbf{Order} &
 & \textbf{Score} & \textbf{Order} \\
\hline
$A'$  & $0.35$ & 2 & $0.4$ & 1 & 1.5 & 0.3708 & 2\\
\hline
$B$  & $0.25$ & 1 & $0.55$ & 2 & 1.5 & 0.3431 & 1\\
\hline
$C$  & $0.7$& 3 & $0.7$ & 3 & 3 & 0.7318 & 3\\
\hline
\end{tabular}
\label{tab:Cd}
}\\
\begin{flushleft}
Note: Different observations of objects would produce different
ranking lists of objects. In (a), objects $A$, $B$ and $C$ can be
ordered by their values on $x_1$ and $x_2$ respectively. Ranking
aggregation (RankAgg) then produce a comprehensive ordering by
Eq.(\ref{eq:rankagg}). But it fails to distinguish $A$ and $B$ which
have distinguishable observations while RPC can distinguish them.
RPC can also detect the minor ordinal difference between objects. In
(b), $A$ has a different observation from (a), which is denoted as
$A'$. Ranking lists keeps the same for RankAgg while RPC provides a
different ordering.
\end{flushleft}
\end{threeparttable}
\end{table*}

For ranking task, some researchers prefer to aggregate many different ranking lists of the
same set of objects in order to get a ``better'' order.
For example, median rank aggregation
\cite{dwork2001} aggregates different orderings into a median rank with
\begin{equation} \label{eq:rankagg}
 \kappa(i)=\frac{\sum_{j=1}^m \tau_j(i)}{m}, i=1,2,\cdots,n
\end{equation}
where $\tau_j(i)$ is the location of object $i$ in ranking list $\tau_j$,
$\tau_j$ is a permutation of $\{1,2,\cdots,n\}$ and $\kappa$ is the ordering of median
rank aggregation.
However, approaches of ranking aggregation suffers the difficulties of strict monotonicity and smoothness.
Therefore, the ranking list is not very convincing.
What's more, aggregation merely combines the orderings and ignores the information
delivered by numerical observations.

In contrast, RPC is modeled following all the five meta-rules which
infers a reasonable ranking list. Moreover, RPC can detect the
ordinal information embedded in the numerical observations,
illustrated in Fig. \ref{fig:illustrate}. Consider to rank three
objects $A$, $B$ and $C$ in a two-dimensional space in Fig.
\ref{fig:illustrate}. Let their numerical observations on $x_1$ and
$x_2$ be values shown in Table \ref{tab:Cb}. Objects can be ordered
along with $x_1$ and $x_2$ respectively. Median rank aggregation
\cite{dwork2001} produces an ordering which can not distinguish $A$
and $B$ since they are in the paratactic place of the ranking list.
In contrast, the RPC model produce the order $ABC$ where $A$ and $B$
are in a distinguishable order since RPC ranks objects based on
their original observation data. If there is a different observation
for one of objects, a different RPCwould produce a different ranking
list while RankAgg remains the same. In Table \ref{tab:Cd}, a
different observation of object $A$ is obtained, denoted as $A'$. A
different RPC is learned (the pink curve in Fig.
\ref{fig:illustrate}) and gives the order $BA'C$ (the last column of
Table \ref{tab:Cd}) which is different from the order in Table
\ref{tab:Cb}. In summary, RPC is able to capture the ordinal
information contained not only among ranking candidates but also in
the individual observation.

\subsection{Applications} \label{ssec:app}

Unsupervised ranking of multi-attribute observations of objects has a widely applications.
The most significant application is to rank countries, journals and universities.
Taking the journal ranking task for illustration,
there have been many indices to rank journals, such as impact factor (IF) \cite{garfield2006}
and Eigenfactor \cite{bergstrom2008}.
Different indices reflect different aspects of journals and provide different ranking lists
for journals. Thus, how to evaluate journals in a comprehensive way becomes a tough
problem. RPC model is proposed as a new framework to attack the problem
which provides an ordering along the ``ranking skeleton'' of data distribution.
In this paper, we perform ranking tasks with RPCs to produce a comprehensive
evaluation on three open access datasets of countries and journals with the open source
software Scilab (5.4.1 version) on a Ubuntu 12.04 system with 4GB memory.
Due to space limitation, we just list parts of their ranking lists
(the full lists will be available when the paper is published).

\subsubsection{Results on Life Qualities of Countries} \label{ssec:qli}

\begin{table*}[t]\scriptsize
\caption{Part of the ranking list for life qualities of countries.}
\label{tab:qli} \centering
\renewcommand{\arraystretch}{1.3}
\begin{threeparttable}
\begin{tabular}{|c||cccc|cc|cc|}
\hline
\multirow{2}{*}{\textbf{Country}}& \multirow{2}{*}{\textbf{GDP}\tnote{1} }
&\multirow{2}{*}{\textbf{LEB}\tnote{2}} &\multirow{2}{*}{ \textbf{IMR}\tnote{3} }
&\multirow{2}{*}{\textbf{Tuberculosis}\tnote{4}}
&\multicolumn{2}{c|}{\textbf{Elmap}\cite{gorban2010b}} & \multicolumn{2}{c|}{\textbf{RPC}} \\
\cline{6-9}
 & & & & &\textbf{Score} & \textbf{Order} &\textbf{Score} &\textbf{Order} \\
\hline\hline
Luxembourg  &70014  &79.56  &6  &4  & 0.892 &1&1.0000&1
\\
Norway  &47551  &80.29  &3  &3  & 0.647 & 2&0.8720&2
\\
Kuwait  &44947  &77.258 &11 &10 & 0.608 & 3&0.8483&3
\\
Singapore   &41479  &79.627 &12 &2  &0.578 & 4&0.8305&4
\\
United States   &41674  &77.93  &2  &7  & 0.575 & 5&0.8275&5
\\
\vdots  &\vdots &\vdots &\vdots &\vdots &\vdots &\vdots &\vdots&\vdots
\\
Moldova &2362   &67.923 &63 &17 &0.002 & 97 & 0.5139 & 96\\
Vanuatu &3477   &69.257 &37 &31 &0.011 & 96 & 0.5135 & 97\\
Suriname &7234  &68.425 &53 &30 &0.011 & 95 & 0.5133 & 98\\
Morocco &3547   &70.443 &44 &36 &0.002 & 98 & 0.5106 & 99\\
Iraq    &3200   &68.495 &25 &37 &-0.002 & 100 & 0.5032 & 100
\\
\vdots  &\vdots &\vdots &\vdots &\vdots &\vdots &\vdots &\vdots&\vdots
\\
South Africa    &8477   &51.803 &349    &55 &-0.652 & 167&0.0786&167 \\
Sierra Leone    &790    &46.365 &219    &160    & -0.664 &169&0.0541&168 \\
Djibouti    &1964   &54.456 &330    &88 & -0.655 & 168&0.0524&169 \\
Zimbabwe    &538    &41.681 &311    &68 & -0.680 & 170&0.0462&170 \\
Swaziland   &4384   &44.99  &422    &110    &-0.876 & 171&0&171 \\
\hline
$\mathbf{p}_0$ &44713 &81.218 &2 &0 &- & - &  - &  -\\
$\mathbf{p}_1$ &330   &80.4   &2 &0 & - & - & - &  -\\
$\mathbf{p}_2$ &330   &59.7   &33 & 43 &- & - &  - &  -\\
$\mathbf{p}_3$ &1581.824 & 41.68 & 290 & 151 &- & - &  - &  -\\
\hline
\end{tabular}
\begin{tablenotes}
 \item[1] Gross Domestic Product per capita by Purchasing Power Parities,
\textdollar per person;
 \item[2] Life Expectancy at Birth, years;
 \item[3] Infant Mortality Rate (per 1000 born);
 \item[4] Infectious Tuberculosis, new cases per 100,000 of population, estimated.
 \item[5] $\mathbf{p}_j(j=0,1,2,3)$ are control and end points of the RPC.
\end{tablenotes}
\end{threeparttable}
\end{table*}

\begin{figure}[t]
\centering
\includegraphics[width = 8cm, height = 8cm, bb = 0 0 600 600]{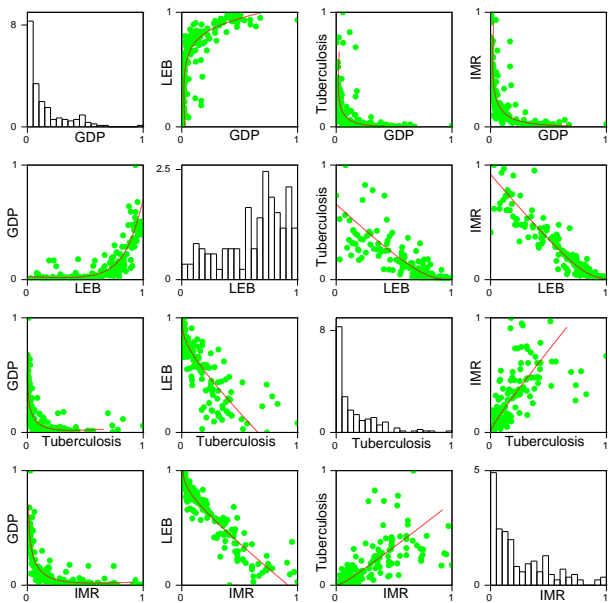}
\centering \caption{Two-dimensional display of data points and RPC
for life qualities of countries. Green points are numerical
observations and red curves are 2-dimensional projection of RPC. }
\label{fig:qli}
\end{figure}

Gorban et al. \cite{gorban2010b} ranked 171 countries by life
qualities of people with data driven from
GAPMINDER\footnote{http://www.gapminder.org/} based on four
indicators as in Example \ref{emp:isotone}. For comparison, we also
use the same four GAPMINDER indicators in \cite{gorban2010b}. The
RPC learned by Algorithm \ref{agrm:RPC} is shown in two-dimensional
visualization in Fig. \ref{fig:qli} and part of the ranking list is
illustrated in Table \ref{tab:qli}.

From Fig. \ref{fig:qli}, RPC portrays the data distributing trends
with different shapes, including linearity and nonlinearity. For
this task, ${\bm\alpha}=[1,1,-1,-1]^T$ for this task just as Example
\ref{emp:isotone}. $\bm\alpha$ also discovers the relationship
between indicators for ranking. GDP is in the same direction with
LEB, but in the opposite direction with IMR and Tuberculosis. In the
beginning, a small amount of GDP increasing brings about tremendous
increasing of LEB and tremendous decreasing of IMR and Tuberculosis.
When GDP exceeds \textdollar14300 (0.2 as normalized value in Fig.
\ref{fig:qli}) per person, increasing GDP does result in little LEB
increase, so does IMR and Tuberculosis decrease. As a matter of
fact, it is hard to improve further LEB, IMR and Tuberculosis when
they are close to the limit of human evolution.

In Table \ref{tab:qli}, control points provided by RPC learning algorithm
(Algorithm \ref{agrm:RPC}) are listed in the bottom.
$\mathbf{p}_i$ in the bottom is given in the original data space.
Although the number of control points are set to two in addition to two end points,
the number actually needed for each indicators is adapted automatically by learning.
From Table \ref{tab:qli}, $\mathbf{p}_0$ and $\mathbf{p}_1$ for IMR and Tuberculosis
overlaps which means that
three points are enough for a B\'{e}zier curve to depicts the skeleton of IMR and Tuberculosis.
Two-dimensional visualizations in Fig. \ref{fig:qli} tally with the statement above.

Gorban et al. \cite{gorban2010b} provided centered scores for
countries, which is similar to the first PCA. But the zero score is
assigned to no country such that no country is taken as the ranking
reference. In addition, rankers would get into trouble to understand
the ranking principle due to unknown parameter size. Therefore, the
ranking list is hard to interpret for human understanding. Compared
with Elmap \cite{gorban2010b}, the presented RPC model follows all
the five meta-rules. With these meta-rules as constraints, it
achieves a better fitting performance in term of Mean Square Error
($90\%$ vs $86\%$ of explained variance). It produces scores in
$[0,1]$ where 0 and 1 are the worst and the best reference
respectively. Luxembourg with the best life quality provides a
developing direction for countries below. Additionally, the RPC
model is interpretable and easy to carry out in practice since there
are just four points to determine the ranking list.
%

\subsubsection{Results on Journal Ranking} \label{ssec:journal}

\begin{table*}[t]\scriptsize
\caption{Part of the ranking list for JCR2012 journals of computer
sciences.} \label{tab:jcr} \centering
\renewcommand{\arraystretch}{1.3}
\begin{tabular}{|c||cc|cc|cc|cc|cc|cc|}
\hline
\multirow{2}{*}{\textbf{Title}} & \multicolumn{2}{c|}{\textbf{Impact Factor (IF)}}
& \multicolumn{2}{c|}{\textbf{5-Year IF}} & \multicolumn{2}{c|}{\textbf{Immediacy Index}}
& \multicolumn{2}{c|}{\textbf{Eigenfactor}} & \multicolumn{2}{c|}{\textbf{Influence Score}}
& \multicolumn{2}{c|}{\textbf{RPC}} \\
\cline{2-13}
 & \textbf{Score} & \textbf{Order} & \textbf{Score} & \textbf{Order}
 & \textbf{Score} & \textbf{Order} & \textbf{Score} & \textbf{Order}
 & \textbf{Score} & \textbf{Order} & \textbf{Score} & \textbf{Order} \\
\hline
\hline
IEEE T PATTERN ANAL &4.795 & 7 &6.144 & 5 &0.625 & 26 &0.05237 & 3 &3.235 & 6 &1.0000 &1 \\

ENTERP INF SYST UK &9.256 & 1 &4.771 & 10 &2.682 & 2 &0.00173 & 230 &0.907 & 86 &0.9505 &2 \\

J STAT SOFTW    &4.910 & 4 &5.907 & 6 &0.753 & 18 &0.01744 & 20 &3.314 & 4 &0.9162&3 \\

MIS QUART &4.659 & 8 &7.474 & 2 &0.705 & 21 &0.01036 & 49 &3.077 & 7 &0.9105 &4\\

ACM COMPUT SURV &3.543 & 21 &7.854 & 1 &0.421 & 56 &0.00640 & 80 &4.097 & 1 &0.9092 &5\\

$\vdots$&$\vdots$ &$\vdots$&$\vdots$ &$\vdots$ &$\vdots$ &$\vdots$ &$\vdots$
&$\vdots$ &$\vdots$ &$\vdots$ &$\vdots$ &$\vdots$\\

DECIS SUPPORT SYST&2.201 & 51 &3.037 & 43 &0.196 & 169 &0.00994 & 52 &0.864 &93 &0.4701&65\\

COMPUT STAT DATA AN&1.304 & 156 &1.449 & 180 &0.415 & 61 &0.02601 & 11 &0.918 & 83 &0.4665&66\\

IEEE T KNOWL DATA EN&1.892 & 82 &2.426 & 72 &0.217 & 152 &0.01256 & 37 &1.129 & 55 &0.4616&67\\

MACH LEARN & 1.467 & 133 & 2.143 & 96 & 0.373 & 70 & 0.00638 & 81 & 1.528 & 20 & 0.4490 & 68\\

IEEE T SYST MAN CY A & 2.183 & 53 & 2.44 & 68 & 0.465 & 46 & 0.00728 & 69 & 0.767 & 111 & 0.4466 & 69\\

$\vdots$&$\vdots$ &$\vdots$&$\vdots$ &$\vdots$ &$\vdots$ &$\vdots$ &$\vdots$
&$\vdots$ &$\vdots$ &$\vdots$ &$\vdots$ &$\vdots$\\
\hline
\end{tabular}
\end{table*}

\begin{figure}[!htbp]
\centering
\includegraphics[width = 8cm, height = 8cm, bb = 0 0 604 604]{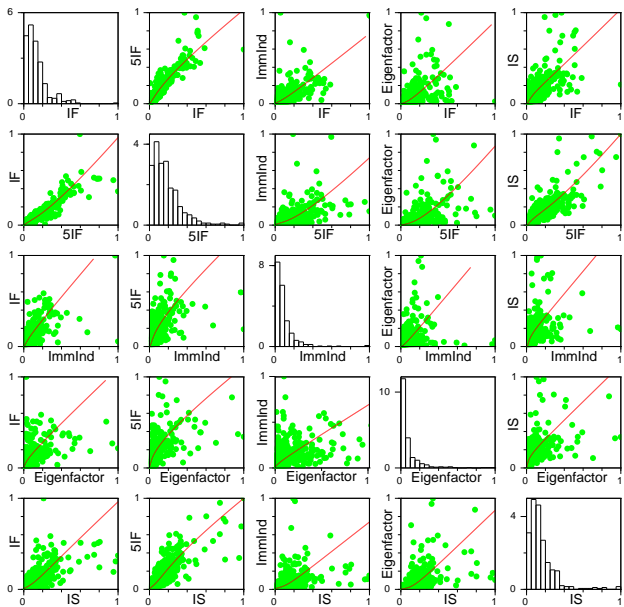}
\centering
\caption{Two-dimensional display of data points and RPC for JCR2012.
Green points are numerical observations and red curves are 2-dimensional projection of RPC.
(IF:Impact Factor, 5IF:5-Year IF, ImmInd:Immediacy Index, IS: Influence Score)}
\label{fig:jcr}
\end{figure}

We also apply RPC model to rank journals with data accessable from the Web of
Knowledge\footnote{http://wokinfo.com/} which is affiliated to Thomson Reuters.
Thomson Reuters publishes annually Journal Citation Reports (JCR) which provide information
about academic journals in the sciences and social sciences.
JCR2012 reports citation information with indicators of Impact Factor, 5-year Impact
Factor, Immediacy Index, Eigenfactor Score, and Article Influence Score.
After journals with data missing are removed from the data table (58 out of 451),
RPC model tries to provide a comprehensive ranking list of journals in the categories of
computer science: artificial intelligence, cybernetics, information systems,
interdisciplinary applications, software engineering, theory and methods.
Table \ref{tab:jcr} illustrates the ranking list of journals produced by RPC model based on JCR2012.
Two-dimensional visualization of the RPC is shown in Fig. \ref{fig:jcr}.

For this ranking task, a journal will rank higher with a higher value for each indicator,
that is ${\bm\alpha}=[1,1,1,1]$.
Among all the indicators here, 5-year Impact Factor shows almost a linear relationship with the others.
But Eigenfactor presents no clear relationship which means that it is calculated in a very different
way from the other indicator. Actually, Eigenfactor works like PageRank \cite{page1998} while
the others take frequency count.

From Table \ref{tab:jcr}, IEEE Transactions on Knowledge and Data
Engineering (TKDE) is ranked in a higher place than IEEE Transactions on
Systems, Man, and Cybernetics-Part A (SMCA) although SMCA has a
higher IF (2.183) than TKDE (1.892). The lower influence score
(0.767) of SMCA brings it down the ranking list (vs. 1.129 for
TKDE). Therefore, TKDE gets a higher comprehensive evaluating score
and wins a higher ranking place in the ranking list. This means that
one indicator does not tell the whole story of ranking lists. RPC
produces a ranking list for journals taking account several
indicators of different aspects.

\section{Conclusions} \label{sec:conclu}

Ranking and its tools have and will have an increasing impact on the
behavior of human, either positively or negatively. However, those
ranking activities are still facing many challenges which have
greatly restrained to the rational design and utilization of ranking
tools. Generally, ranking in practice is an unsupervised task which
encounters a critical challenge that there is no ground truth to
evaluate the provided lists. PageRank \cite{page1998} is an
effective unsupervised ranking model for ranking candidates with a
link-structure. However, it does not work for numerical observations
on multiple attributes of objects.

It is well known that domain knowledge can always improve the data
mining performance. We try to attack unsupervised ranking problems
by domain knowledge about ranking. Motivated by
\cite{hubg2009,kotlow2013}, five meta-rules as ranking knowledge are
presented and are regarded as constraints to ranking models. They
are scale and translation invariance, strict monotonicity,
linear/nonlinear capacities, smoothness and explicitness of
parameter size. They can also be capable of assessing the ranking
performance of different models. Enlightened by
\cite{hubg1998,gorban2010b}, we propose a ranking model with a
principal curve which is parametrically formulated with a cubic
B\'{e}zier curve by restricting control points in the interior of 
the hypercube $[0,1]^d$. Control points are learned from the data
distribution without human interventions. Applications in life
qualities of countries and journals of computer sciences show that
the proposed RPC model can produce reasonable ranking lists.

From an application view points, there are many indicators for ranking objects.
RPC can also be used to do feature selection which is one part of our future works.

\appendices

\section{Principal Curves} \label{app:pc}

\setcounter{equation}{0}
\renewcommand{\theequation}{\ref{app:pc}-\arabic{equation}}

Given a dataset
$\mathbf{X}=(\mathbf{x_1},\mathbf{x_2},\cdots,\mathbf{x_n})$, $\mathbf{x}_i\in{\mathbf{R}^d}$,
a principal curve summarizes the data with a smooth curve
instead of a straight line in the first PCA
\begin{equation} \label{eq:pcmodel}
\mathbf{x} = \mathbf{f}(s) + {\bm\varepsilon}
\end{equation}
where $\mathbf{f}(s)=(f_1(s),f_2(s),\cdots,f_d(s))\in{\mathbf{R}^d}$
and $s\in{\mathbf{R}}$.
The principal curve $\mathbf{f}$ was originally defined by Hastie and Stuetzle \cite{hastie1989}
as a smooth ($C^\infty$) unit-speed ($\|\mathbf{f}''\|^2=1$)
one-dimensional manifold in $\mathbf{R}^d$ satisfying the self-consistence condition
\begin{equation*}
 \mathbf{f}(s) =
E\left(\left.\mathbf{x}\right|s_{\mathbf{f}}(\mathbf{x})=s\right)
\end{equation*}
where $s=s_{\mathbf{f}}(\mathbf{x})\in{\mathbf{R}}$ is the largest value
so that $\mathbf{f}(s)$ has the minimum distance from $\mathbf{x}$.
Mathematically, $s_{\mathbf{f}}(\mathbf{x})$ is formulated as \cite{hastie1989}
\begin{equation} \label{eq:lambdahs}
 s_{\mathbf{f}}(\mathbf{x})=
\sup\left\{s:\|\mathbf{x}-\mathbf{f}(s)\|=
\inf_s \|\mathbf{x}-\mathbf{f}(s)\|\right\}.
\end{equation}
In other words, a curve $\mathbf{f}:\mathbf{R}\mapsto\mathbf{R}^d$ is called a \textit{principal
curve} if it minimizes the expected squared distance between $\mathbf{x}$ and $\mathbf{f}$
which is denoted by \cite{kegl2000}
\begin{equation} \label{eq:pcobjective}
 J(\mathbf{f}) = E\left(\inf_s\|\mathbf{x}-\mathbf{f}(s)\|^2\right)
=E\|\mathbf{x}-\mathbf{f}(s_{\mathbf{f}}(\mathbf{x}))\|^2.
\end{equation}

As an one-dimensional principal manifold, the principal curve has a wide applications
(e.g. \cite{zhang2011tnn}) due to its simpleness.
Following Hastie and Stuetzle \cite{hastie1989}, researchers afterwards have proposed
a variety of principal curve definitions and learning algorithms to perform different
tasks \cite{brpc1992,kegl2000,einbeck2005,gorban2010}.
But most of them tried to first approximate the principal curve first with a polyline
\cite{kegl2000} and then smooth it to meet the requirement for smoothness \cite{hastie1989} of
the principal curve. Therefore, the expression of the principal curve is not
explicit and results in a `black-box' which is hard to interpret.
The other definitions of principal curves
\cite{tibsh1992,delicado2001} employed Gaussian mixture model to generally
formulate the principal curve which brings model bias and makes interpretation even harder.
When the principal curve is used to perform a ranking task, it should be modeled to be a
`white-box' which can be well interpreted for its provided ranking lists.

\section{Proof of Theorem \ref{thm:monoinverse}}
\label{app:monoproof}

\setcounter{equation}{0}
\renewcommand{\theequation}{\ref{app:monoproof}-\arabic{equation}}

If $\nabla\varphi(\mathbf{x})\succ\mathbf{0}$, $\varphi$ is strictly monotone by
Theorem \ref{thm:monovecsuff}.
Regarding that the ranking candidates is totally ordered, there is a one-to-one correspondence
between ranking items in $\mathbf{R}^d$ and $\mathbf{rang}\varphi$.
Otherwise, $s=\varphi(\mathbf{x}_0)$ and $s=\varphi(\mathbf{x}_0+\triangle\mathbf{x})$ both hold
for some $\mathbf{x}_0\in\mathbf{dom}\varphi$.
In this case, $\left.\nabla\varphi(\mathbf{x})\right|_{\mathbf{x}=\mathbf{x}_0}=\mathbf{0}$
which contradicts the assumption $\nabla\varphi(\mathbf{x})\succ\mathbf{0}$ holds for all
$\mathbf{x}\in\mathbf{dom}\varphi$.

By Lemma \ref{lem:mono} and the one-to-one correspondence, there exists an inverse mapping
$\mathbf{f}:\mathbf{rang}\varphi\mapsto\mathbf{dom}\varphi$ such that $\mathbf{x}=\mathbf{f}(s)$.
By strict monotonicity (Eq.(\ref{def:ordkeep})) and the one-to-one correspondence, we have
\begin{equation}
 \mathbf{x}_1\preceq\mathbf{x}_2,\quad\mathbf{x}_1\neq\mathbf{x}_2 \Longleftrightarrow s_1<s_2
\end{equation}
Thus, $\nabla\mathbf{f}(s)\succ\mathbf{0}$ holds for $s\in\mathbf{rang}\varphi$. $\hfill\square$

\section{Proof of RPC Existence (Theorem \ref{thm:existence})} \label{app:existproof}

\setcounter{equation}{0}
\renewcommand{\theequation}{\ref{app:existproof}-\arabic{equation}}

\textit{Proof.}
Assume $\mathbf{U}=[0,1]$ and $C(\mathbf{U})$ denotes the set of all continuous function
$\mathbf{f}:\mathbf{U}\mapsto[0,1]^d\subseteq{\mathbf{R}^d}$ embracing
all possible observations of $\mathbf{x}$.
The uniform metric is defined as
\begin{equation}
 D(\mathbf{f}, \mathbf{g}) = \sup_{0\leq s\leq 1}
\|{\mathbf{f}(s)-\mathbf{g}(s)}\|, \quad
\forall \mathbf{f},\mathbf{g}\in{C(\mathbf{U})}.
\end{equation}
It is easy to see $(C(\mathbf{U}),D)$ is a complete metric space \cite{boyd2004}.

Let $\mathbf{\varGamma}=\left\{\mathbf{f}(s):\mathbf{f}(s)= \mathbf{PMz},
\mathbf{P}\in{\Theta}\right\}
\subseteq{C(\mathbf{U})}$, where $\Theta\in[0,1]^4$ is the convex hull of
$\mathbf{x}$.
With the Frobenius norm, $\Theta$ is a sequentially compact set so that
for any given sequence in $\Theta$ there exists a subsequence $\mathbf{P}^{(t)}$ converging
uniformly to an $\mathbf{P}^*\in[0,1]^d$ \cite{boyd2004} with
\begin{equation}
 \| \mathbf{P}^{(t)}-\mathbf{P}^*\|_F \hspace{0.2cm}\rightarrow 0
\end{equation}
Let $\mathbf{p}_0=\frac{1}{2}(\mathbf{1}-\bm\alpha)$ and
$\mathbf{p}_3=\frac{1}{2}(\mathbf{1}+\bm\alpha)$.
Then we have a sequence $\mathbf{f}^{(t)}(s)$ converging uniformly to
$\mathbf{f}^*(s)$:
\begin{eqnarray}
\hspace{-1cm}& &D\left(\mathbf{f}^{(t)}(s),\mathbf{f}^*(s)\right)
=\sup_{0\leq s\leq 1}\|{\mathbf{f}^{(t)}(s)-\mathbf{f}^*(s)}\| \\
\hspace{-1cm}&\leq&\sup_{0\leq s\leq 1}\| \mathbf{P}^{(t)}-\mathbf{P}^*\|_F\|\mathbf{Mz}\| \\
\hspace{-1cm}&=&\| \mathbf{P}^{(t)}-\mathbf{P}^*\|_F \hspace{0.2cm}\rightarrow 0
\end{eqnarray}
where $\|\mathbf{Mz}\|=1$.
By Proposition \ref{prop:mono}, $\mathbf{f}^{(t)}(s)$ is a curve sequence of
strictly monotonicity and converges to $\mathbf{f}^*(s)$.

Assuming the converging sequence $\mathbf{f}^{(t)}(s)$ makes $J(\mathbf{P}^{(t)})
\geq{J(\mathbf{P}^*)}$
for fixed $\mathbf{x}\in\mathbf{R}^d$,
\begin{eqnarray}
\hspace{-1cm} & &J(\mathbf{P}^{(t)})-J(\mathbf{P}^*) \nonumber \\
\hspace{-1cm} &=& \|\mathbf{x}-\mathbf{f}^{(t)}(s)\|^2-\|\mathbf{x}-\mathbf{f}^*(s) \|^2 \\
\hspace{-1cm}&\leq& \left(\| \mathbf{x}-\mathbf{f}^{(t)}(s)\|+\| \mathbf{x}-\mathbf{f}^*(s)\|\right)
      \| \mathbf{f}^{(t)}(s)-\mathbf{f}^*(s)\|  \\
\hspace{-1cm}& &\rightarrow 0
\end{eqnarray}
and therefore
\begin{equation}
 E\left(J(\mathbf{P}^{(t)})-J(\mathbf{P}^*)\right)\hspace{0.2cm}\rightarrow 0.
\end{equation}
Finally, we complete the proof.
$\hfill\square$

\section{Proof of Convergence (Proposition \ref{prop:converge})} \label{app:propproof}

\setcounter{equation}{0}
\renewcommand{\theequation}{\ref{app:propproof}-\arabic{equation}}

\textit{Proof:}
First of all, $\mathbf{P}^{(t)}$ generated by Richardson method has been proved to converge
\cite{Richard1910}.  Assume $\mathbf{P}^{(t)}\rightarrow{\mathbf{P}^*}$, and
$\mathbf{s}^{(t)}$ and $\mathbf{s}^*$ are the corresponding
score vectors calculated by Eq.(\ref{eq:updateT}). Note that the item
$\mathbf{P}^{(t+1)}-\mathbf{P}^{(t)}$ is in the descending direction of $J$ in
Eq.(\ref{eq:numiterP}). So we get that
\begin{equation}
J(\mathbf{P}^{(t)}, \mathbf{s}^{(t)}) \geq J(\mathbf{P}^{(t+1)}, \mathbf{s}^{(t)}).
\end{equation}
Then with the control points $\mathbf{P}^{(t+1)}$, $\mathbf{s}^{(t+1)}$ minimizes the summed
orthogonal distance
\begin{equation}
J(\mathbf{P}^{(t+1)}, \mathbf{s}^{(t)}) \geq J(\mathbf{P}^{(t+1)}, \mathbf{s}^{(t+1)}).
\end{equation}
Thus we get
\begin{equation}
 J(\mathbf{P}^{(t)}, \mathbf{s}^{(t)}) \geq J(\mathbf{P}^{(t+1)}, \mathbf{s}^{(t+1)}).
\end{equation}
Finally, by Theorem \ref{thm:existence} the sequence $\{J(\mathbf{P}^{(t)},\mathbf{s}^{(t)})\}$
converges to its infimum $\{J(\mathbf{P}^*,\mathbf{s}^*)\}$ as $s\rightarrow\infty$.
$\hfill\square$


\section*{Acknowledgement}
The authors appreciate very much the advice from the machine learning crew in NLPR.
This work is supported in part by NSFC (No. 61273196) for C.-G. Li and B.-G. Hu,
and NSFC (No. 61271430 and No. 61332017) for X. Mei.

\ifCLASSOPTIONcaptionsoff
  \newpage
\fi




\begin{thebibliography}{00}

\bibitem{lihang2011}
H. Li, ``A Short Introduction to Learning to Rank'',
\emph{IEICE Trans. Inf. Syst.}, vol. E94-D, no. 10, pp. 1-9, 2011.

\bibitem{page1998}
S. Brin and L. Page,
``The Anatomy of a Large-Scale Hypertextual Web Search Engine'',
\emph{Computer Networks}, vol. 30, no. 1-7, pp. 107-117, 1998.

\bibitem{zhou2004}
D. Zhou, J. Weston, A. Gretton, O. Bousquet, and B. Sch\"{o}lkopf,
``Ranking on Data Manifolds'',
\emph{Advances in Neural Information Processing Systems 16},
S. Thrun, L. Saul, and B. Sch\"{o}lkopf, eds., MIT Press, 2004.

\bibitem{cai2011}
B. Xu, J. Bu, C. Chen, D. Cai, X. He, W. Liu, and J. Luo,
``Efficient manifold ranking for image retrieval'',
\emph{Proc. 34th Int'l ACM SIGIR Conf. Research
and Development in Information Retrieval}, pp. 525-534, 2011.

\bibitem{bishop2006}
C. Bishop, \emph{Pattern Recognition and Machine Learning}, New York: Springer, 2006.

\bibitem{guyon2003}
I. Guyon and A. Elisseeff,
``An Introduction to Variable and Feature Selection'',
\emph{J. Mach. Learn. Res.}, vol. 3, pp. 1157-1182, 2003.

\bibitem{small2009}
A. Klementiev, D. Roth, K. Small, and I. Titov,
``Unsupervised Rank Aggregation with Domain-Specific Expertise'',
\emph{Proc. 20th Int'l Joint Conf. Artifical Intell.}, pp. 1101-1106, 2009.

\bibitem{gorban2010b}
A.Y. Zinovyev and A.N. Gorban, ``Nonlinear Quality of Life Index''[EB/OL],
New York, \url{http://arxiv.org/abs/1008.4063}, 2010.

\bibitem{vasuki2006}
A. Vasuki, ``A Review of Vector Quantization Techiniques'',
\emph{IEEE Potentials}, vol. 25, no. 4, pp. 39-47, 2006.

\bibitem{hastie1989}
T. Hastie and W. Stuetzle, ``Principal Curves'',
\emph{J. Amer. Stat. Assoc.}, vol. 84, no. 406, pp. 502-516, 1989.

\bibitem{kegl2000}
B. K\'{e}gl, A. Krzy\.{z}ak, T. Linder, and K. Zeger,
``Learning and Design of Principal Curves'',
\emph{IEEE Trans. Pattern Anal. Mach. Intell.}, vol. 22, no. 3, pp. 281-297, 2000.

\bibitem{gould1981}
P. Gould, ``Letting the Data Speak for Themselves'',
\emph{Assoc. Amer. Geog. USA}, vol. 71, no. 2, 1981.

\bibitem{hubg2009}
B.-G. Hu, H.B. Qu, Y. Wang, and S.H. Yang,
``A Generalized-Constraint Neural Network model: Associating Partially Known Relationships for
Nonlinear Regression'', \emph{Inf. Sci.}, vol. 179, pp. 1929-1943, 2009.

\bibitem{hubg1998}
B.-G. Hu, G.K.I Mann, and R.G. Gosine,
``Control curve design for nonlinear (or fuzzy) proportional actions using spline-based
functions'', \emph{Automatica}, vol. 34, no. 9, pp. 1125-1133, 1998.

\bibitem{daniels2010tnn}
H. Daniels and M. Velikova,
``Monotone and Partially Monotone Neural Networks'',
\emph{IEEE Trans. Neural Networks}, vol. 21, no. 6, pp. 906-917, 2010.

\bibitem{kotlow2013}
W. Kot\l{}owski and R. S\l{}owi\'{n}ski,
``On Nonparametric Ordinal Classification with Monotonicity Constraints'',
\emph{IEEE Trans. Knowl. Data Engineering}, vol. 25, no. 11, pp. 2576-2589, 2013.

\bibitem{pei2014tkde}
Y. Zhang, W. Zhang, J. Pei, X. Lin, Q. Lin, and A. Li,
``Consensus-Based Ranking of Multivalued Objects: A Generalized Borda Count Approach'',
\emph{IEEE Trans. Knowl. Data Engineering}, vol. 26, no. 1, pp. 83-96, 2014.

\bibitem{cheng2013}
X.Q. Cheng, P. Du, J.F. Guo, X.F. Zhu, and Y.X. Chen,
``Ranking on Data Manifold with Sink Points'',
\emph{IEEE Trans. Knowl. Data Engineering}, vol. 25, no. 1, pp. 177-191, 2013.

\bibitem{gorban2010}
A.N. Gorban and A.Y. Zinovyev,
``Chapter 2: Principal Graphs and Manifolds'',
\emph{Handbook of Research on Machine Learning Applications and
Trends: Algorithms, Methods, and Techiniques},
E.S. Olivas, J.D.M. Guerrero, M.M. Sober, J.R.M. Benedito, A.J.S. L\'{o}pez, eds.,
New York: Inf. Sci. Ref., vol. 1, pp. 28-59, 2010.

\bibitem{pastva1998}
T.A. Pastva, ``B\'{e}zier Curve Fitting'',
master's thesis, Naval Postgraduate School, 1998.

\bibitem{boyd2004}
S. Boyd and L. Vandenberghe,
\emph{Convex Optimization},
New York: Camb. Univ. Press, 2004.

\bibitem{priestley2002}
H.A. Priestley,
``Chapter 2: Ordered Sets and Complete Lattices-a Primer for Computer Science'',
\emph{Algebraic and Coalgebraic Methods in the Mathematics of Program Construction},
R. Backhouse, R. Crole, J. Jibbons, eds., LNCS 2297, pp. 21-78, 2002.

\bibitem{fitzpatrick2006}
P.M. Fitzpatrick, \emph{Advanced Calculus},
CA: Thomson Brooks/Cole, 2006.

\bibitem{cambini2003}
A. Cambibi, D.T. Luc, and L. Martein.
``Order-Preserving Transformations and Applications'',
\emph{J. Optimization Theory Applications}, vol. 118, no. 2, pp. 275-293, 2003.

\bibitem{anderson2003}
T.W. Anderson, \emph{An Introduction to Multivariate Statistical Analysis},
New Jersey: John Wiley \& Sons, Inc. 2003.

\bibitem{brpc1992}
J.D. Banfield and A.E. Raftery,
``Ice Floe Identification in Satellite Images Using Mathematical
Morphology and Clustering about Pincipal Curves'',
\emph{J. Amer. Stat. Assoc.}, vol. 87, no. 417, pp. 7-16, 1992.

\bibitem{delicado2001}
P.~Delicado,
``Another look at principal curves and surfaces'',
\emph{J. Multivariate Anal.}, vol. 77, no. 1, pp. 84-116, 2001.

\bibitem{chang2001}
K. Chang and J. Ghosh,
``A Unified Model for Probabilistic Principal Surfaces'',
\emph{IEEE Trans. Pattern Anal. Mach. Intell.}, vol. 23, no. 1, pp. 22-41, 2001.

\bibitem{einbeck2005}
J. Einbeck, G. Tutz, and L. Evers, ``Local Principal Curves'',
\emph{Stat. and Comput.}, vol. 15, no. 4, pp. 301-313, 2005

\bibitem{tibsh1992}
R. Tibshirani, ``Principal Curves Revisited'',
\emph{Stat. and Comput.}, vol. 2, no. 4, pp. 183-190, 1992.

\bibitem{farin1997}
G. Farin,
\emph{Curves and Surfaces for Computer Aided Geometric Design (4th Edition)},
California: Acad. Press, Inc., 1997.

\bibitem{jenkins1970}
M. A. Jenkins and J. F. Traub,
``A Three-Stage Algorithm for Real Polynomials Using Quadratic Iteration'',
\emph{SIAM J. Numer. Anal.}, vol. 7, no. 44, pp. 545–566, 1970.

\bibitem{bazaraa2006}
M.S. Bazaraa, H.D. Sherali, and C.M. Shetty,
\emph{Nonlinear Programming: Theory and Algorithms},
New Jersey, Hoboken: John Wiley \& Sons, Inc., 2006.

\bibitem{dwork2001}
C. Dwork, R. Kumar, M. Naor, and D. Sivakumar,
``Rank Aggregation Methods for the Web'',
\emph{Proc. 10th Int'l Conf. World Wide Web}, pp. 613-622, 2001.

\bibitem{roger1985}
R.A. Roger and C.R. Johnson,
\emph{Matrix Analysis},
New York: Camb. Univ. Press, 1985.

\bibitem{zhang2011tnn}
J.P. Zhang, X.D. Wang, U.Kruger, F.Y. Wang,
``Principal Curve Algorithms for Partitioning High-Dimensional Data Spaces'',
\emph{IEEE Trans. Neural Networks}, vol. 22, no. 3, pp. 367-380, 2011.

\bibitem{Richard1910}
L.F. Richardson,
``The approximate arithmetical solution by finite differences of physical problems
involving differential equations, with an application to the stresses in a masonry dam'',
\emph{Philos. Trans. Roy. Soc. London Ser. A}, vol. 210, pp. 307-357, 1910.

\bibitem{golub1996}
G. H. Golub and C. F. van Loan, \emph{Matrix Computations, 3rd ed.},
Baltimore, MD: Johns Hopkins, 1996.

\bibitem{garfield2006}
E. Garfield, ``The History and Meaning of the Journal Impact Factor'',
\emph{J. Amer. Med. Assoc.}, vol. 295, no. 1, pp. 90-93, 2006.

\bibitem{bergstrom2008}
C.T. Bergstrom, J.D. West, M.A. Wiseman,``The Eigenfactor Metrics'',
\emph{J. of Neuroscience}, vol. 28, no. 45, pp. 11433-11434, 2008.

\end{thebibliography}
%

  \vfill\newpage

\end{document}